\documentclass[journal,twoside,web]{ieeecolor}
\usepackage{comment}

\usepackage{amsmath,amssymb,amsfonts}
\usepackage{algorithmic}
\usepackage{graphicx}
\usepackage{textcomp}
\usepackage{snapshot}

\usepackage{silence}
\ErrorFilter{LaTeX Error: Option clash}

\def\logoname{LOGO-generic-web}
\definecolor{subsectioncolor}{rgb}{0,0.541,0.855}
\def\journalname{\MakeUppercase{XXXXX XXXXX XXXXX} }

\usepackage[utf8]{inputenc}
\usepackage[T1]{fontenc}
\usepackage{tabularx}
\usepackage{soul}
\usepackage{multirow}
\usepackage{booktabs}
\usepackage{mathtools, nccmath}

\makeatletter
\let\NAT@parse\undefined
\makeatother
\usepackage[unicode=true,
bookmarksopen={true},
pdffitwindow=true,
colorlinks=true,
linkcolor=subsectioncolor,
citecolor=subsectioncolor,
urlcolor=subsectioncolor,
hyperfootnotes=true,
pdfstartview={FitH},
pdfpagemode= UseNone]{hyperref}

\usepackage[capitalise]{cleveref} 
\definecolor{darkgray}{rgb}{0.435294,0.435294,0.396078}
\definecolor{darkblue}{rgb}{0.121569,0.200000,0.513725}
\definecolor{oricolor}{rgb}{0.741176,0.717647,0.419608}
\definecolor{multisocolor}{rgb}{0.741176,0.717647,0.419608}
\definecolor{ikcolor}{rgb}{1.000000,0.388235,0.278431}
\definecolor{insolecolor}{rgb}{0.576471,0.439216,0.858824}
\definecolor{idcolor}{rgb}{0.000000,0.501961,0.501961}
\definecolor{socolor}{rgb}{0.121569,0.200000,0.396078}
\definecolor{darkgreen}{rgb}{0.121569,0.474510,0.121569}
\definecolor{verylightgreengray}{rgb}{0.960784,0.980392,0.960784}
\definecolor{lightgreengray}{rgb}{0.860784,0.880392,0.860784}
\definecolor{lightgray}{rgb}{0.860784,0.860784,0.860784}

\definecolor{darkgreen}{rgb}{0.0, 0.5, 0.13}
\definecolor{carmine}{rgb}{0.59, 0.0, 0.09}
\definecolor{cobalt}{rgb}{0.0, 0.28, 0.67}
\definecolor{cardinal}{rgb}{0.77, 0.12, 0.23}

\newcommand{\ruoli}[1]{\textcolor{cobalt}{\textbf{Ruoli:} #1}}

\newcommand{\frederico}[1]{\textcolor{darkgreen}{\textbf{Frederico:} #1}}

\usepackage{ifthen}

\newboolean{removeSTfromtext}
\setboolean{removeSTfromtext}{true} 

\ifthenelse{\boolean{removeSTfromtext}}
  {\newcommand{\ruolist}[1]{}
  \newcommand{\fredericost}[1]{}
  }
  {\newcommand{\ruolist}[1]{\ruoli{\st{#1}}}
  \newcommand{\fredericost}[1]{\frederico{\st{#1}}} }

    \def \variableEventA {Joint angles produced}
    \def \variableEventB {Read joint angles from buffer}
    
    \def \variableEventD { Found wrenches in buffer }
    
    \def \variableEventF { Immediately before \ac{ID}}
    \def \variableEventG { Joint torques calculated}
    \def \variableEventH { Received synchronized joint angles and torques}
    \def \variableEventI { Immediately before \ac{SO} computation}
    \def \variableEventJ { Muscle activations calculated}

    \def \artrunkmovementname {trunk flexion}
    \def \arelevationmovementname {arm elevation}
    \def \arcurlmovementname{elbow flexion}

\usepackage{collcell}
\makeatletter
\newcolumntype{Z}{>{\collectcell\@gobble}c<{\endcollectcell}@{}}
\newcolumntype{C}{>{\centering\arraybackslash}X}
\makeatother
\newcommand{\clabel}[1]{\hypertarget{#1}{}\label{#1}}

\usepackage{CJKutf8}
\makeatletter
\let\MYcaption\@makecaption
\makeatother

\usepackage[font=footnotesize]{subcaption}

\makeatletter
\let\@makecaption\MYcaption
\makeatother

\usepackage{tikz}
\usetikzlibrary{positioning,fit,calc,arrows.meta, backgrounds,shadows.blur, bending}
\usetikzlibrary{shapes,positioning,decorations.markings}
\usepackage{mathtools}

\usepackage{ifthen}
\usepackage{tikz}
\usepackage[outline]{contour} 
\usetikzlibrary{calc}
\usetikzlibrary{angles,quotes} 
\usetikzlibrary{patterns}
\tikzset{>=latex} 
\contourlength{1.4pt}

\colorlet{xcol}{blue!70!black}
\colorlet{vcol}{green!60!black}
\colorlet{myred}{red!65!black}
\colorlet{acol}{red!50!blue!80!black!80}
\tikzstyle{mass}=[line width=0.6,red!30!black,fill=red!40!black!10,rounded corners=1,
                  top color=red!40!black!20,bottom color=red!40!black!10,shading angle=20]
\tikzstyle{velocity}=[->,vcol,very thick,line cap=round]
\tikzstyle{ground}=[preaction={fill,top color=blue!70!black!10,bottom color=blue!70!black!5,shading angle=20},
                    fill,pattern color=blue!30!black,pattern=north east lines,draw=none,minimum width=0.3,minimum height=0.6]
\tikzstyle{limb}=[thick,line cap=round]
\tikzset{armscratch/.pic = {
            \draw[limb,blue] (N)++(-95:0.03) to[out=-105,in=90] ++(-0.1,-0.2*\H) to[out=-90,in=180] ++(-0.1,0.4*\H);
}}
\tikzset{rarmbent/.pic={
            \draw[limb,black] (N)++(-100:0.1) to[out=-45,in=165] ++(0.2,-0.2*\H) to[out=0,in=180] ++(0.3,0.04*\H);

}}
\tikzset{larmbent/.pic={
            \draw[limb] (N)++(-105:0.1) to[out=150,in=160] ++(0.2,-0.19*\H) to[out=0,in=180] ++(0.3,0.04*\H);

}}
\tikzset{rarmn/.pic={
\draw[limb] (N)++(-95:0.03) to[out=-105,in=90] ++(-0.1,-0.4*\H);
}}
\tikzset{larmn/.pic={
\draw[limb] (N)++(-85:0.03) to[out=-80,in=90] ++(0.1,-0.4*\H);
}}

\tikzset{
      pics/human/.style n args={7}{
      code =
      {
        \begin{scope}[rotate=#7]
            \draw[thick] (\x,\H) circle(0.3) coordinate (H);
            \draw[thick] (H)++(-90:0.3) coordinate (N) to[out=-92,in=92]++ (0,-0.40*\H) coordinate (P);

                    \end{scope}

    \begin{scope}[rotate=#1]
        \draw[limb,myred] (P)  -- ++(0,-0.5) coordinate (RK);
        \begin{scope}[rotate=#2]
          \draw[limb,myred] (RK) -- ++(0,-0.45) coordinate (RH);
          \draw[limb,myred,rotate=#3] (RH)-- ++(0.14,0);
        \end{scope}
    \end{scope}

    \draw[thick,myred] (P) circle(0.02);
    \draw[thick,myred] (RK) circle(0.02);
    \draw[thick,myred] (RH) circle(0.02);

    \begin{scope}[rotate=#4]
        \draw[limb] (P)  -- ++(0,-0.5) coordinate (LK);
        \begin{scope}[rotate=#5]
          \draw[limb] (LK) -- ++(0,-0.45) coordinate (LH);
          \draw[limb,rotate=#6] (LH)-- ++(0.14,0);
        \end{scope}
    \end{scope}

    \draw[thick] (P) circle(0.02);
    \draw[thick] (LK) circle(0.02);
    \draw[thick] (LH) circle(0.02);

  }}  }

\tikzset{
    humanwalk/.pic={
        \pic{human={20}{-10}{0}{-25}{-15}{0}{-2}};
                    \pic{rarmn};
            \pic{larmn};
    }
}

\tikzset{
    humanclimb/.pic= {
        \pic{human={30}{-40}{20}{-15}{-20}{-10}{-2}};
                    \pic{rarmn};
            \pic{larmn};
    }
}

\tikzset{
    humanstep/.pic= {
        \pic{human={50}{-60}{10}{-15}{-15}{-10}{-2}};
                    \pic{rarmn};
            \pic{larmn};
    }
}

\def\walkingHuman{

\begin{tikzpicture}
  \def\x{-0.1*\W} 
  \coordinate (O) at (0,0);

  \draw[thin,brown!40!black,fill=brown!80!black,rounded corners=0.5]
    (-0.8*\L,0) --++ (\L,0) |-++ (-\L,\T) -- cycle;

\pic[local bounding box=h1]{humanwalk};

\end{tikzpicture}

}

\def\squatHuman{

\begin{tikzpicture}
\def\x{0} 
  \draw[thin,brown!40!black,fill=brown!80!black,rounded corners=0.5]
    (-0.8*\L,0) --++ (\L,0) |-++ (-\L,\T) -- cycle;

   \draw (-1.0,-.35) pic{human={70}{-120}{45}{65}{-120}{55}{-20}};
            \pic{rarmbent};

            \pic{larmbent};

\end{tikzpicture}

}

\tikzset{
      pics/humannoarm/.style n args={7}{
      code =   
      {  
        \begin{scope}[rotate=#7]
            \draw[thick] (\x,\H) circle(0.3) coordinate (H);
            \draw[thick] (H)++(-90:0.30) coordinate (N) to[out=-92,in=92]++ (0,-0.40*\H) coordinate (P);

        \end{scope}
  
    \begin{scope}[rotate=#1]
        \draw[limb,myred] (P)  -- ++(0,-0.5) coordinate (RK);
        \begin{scope}[rotate=#2]
          \draw[limb,myred] (RK) -- ++(0,-0.45) coordinate (RH);
          \draw[limb,myred,rotate=#3] (RH)-- ++(0.14,0);            
        \end{scope}          
    \end{scope}
      
    \draw[thick,myred] (P) circle(0.02);
    \draw[thick,myred] (RK) circle(0.02);
    \draw[thick,myred] (RH) circle(0.02);
    
    \begin{scope}[rotate=#4]
        \draw[limb] (P)  -- ++(0,-0.5) coordinate (LK);
        \begin{scope}[rotate=#5]
          \draw[limb] (LK) -- ++(0,-0.45) coordinate (LH);
          \draw[limb,rotate=#6] (LH)-- ++(0.14,0);            
        \end{scope}          
    \end{scope}
    
    \draw[thick] (P) circle(0.02);
    \draw[thick] (LK) circle(0.02);
    \draw[thick] (LH) circle(0.02);
 
  }}  }

\def\stsHuman{

\begin{tikzpicture}
\def\L{2.2}     
\def\x{0} 
  \draw[thin,brown!40!black,fill=brown!80!black,rounded corners=0.5]
    (-0.8*\L,0) --++ (\L,0) |-++ (-\L,\T) -- cycle;

\draw[thin,gray!40!black,fill=gray!80!black,rounded corners=0.5]
    (-0.8*\L+.1,0.1) --++ (\L/5,0) |-++ (-\L/5,.45) -- cycle;

 \def\x{-0.2} 
   \draw (-1.5,-.43) pic{humannoarm={90}{-90}{0}{87}{-90}{5}{-5}};
            \pic{rarmbent};

            \pic{larmbent};
            
 \def\x{1} 
\draw[thin,gray!40!black,fill=gray!80!black,rounded corners=0.5]
    (-.4,0.1) --++ (\L/5,0) |-++ (-\L/5,.45) -- cycle;
 \draw (-0.8*\L+.1+.8,.05) pic{human={1}{0}{0}{-1}{0}{0}{0}};
                    \pic{rarmn};
            \pic{larmn};

\end{tikzpicture}

}

\def\W{5.2}     
\def\L{1.6}     
\def\T{0.08}    
\def\H{2.2}     

\usepackage{float}

\usepackage{array}
\newcommand{\PreserveBackslash}[1]{\let\temp=\\#1\let\\=\temp}
\newcolumntype{G}[1]{>{\PreserveBackslash\centering}p{#1}}
\usepackage[normalem]{ulem}

\usepackage{siunitx}
\DeclareSIUnit\year{yr}

\newcommand{\age}[1]{age = \SI[round-precision = 2]{#1}{\year}}

\newcommand{\height}[1]{height = \SI[round-precision = 3]{#1}{\meter}}

\newcommand{\weight}[1]{weight = \SI[round-precision = 2]{#1}{\kg}}

\usepackage{textcomp,gensymb}

\usepackage[symbol]{parnotes}
\usepackage{bm}

\tikzset{
      tf/.pic={
            \draw (0.2,-0.2) node[draw, fill=white,text width=0.6cm,minimum height=0cm]                  {TF};
            \draw (0.1,-0.1) node[draw, fill=white,text width=0.6cm,minimum height=0cm]                  {TF};
            \node[draw, fill=white,text width=0.6cm,minimum height=0cm]                  {TF};
      }
    }

\tikzset{
    hyperlink node/.style={
        alias=sourcenode,
        append after command={
            let     \p1 = (sourcenode.north west),
                \p2=(sourcenode.south east),
                \n1={\x2-\x1},
                \n2={\y1-\y2} in
            node [inner sep=0pt, outer sep=0pt,anchor=north west,at=(\p1)] {\hyperlink{#1}{\XeTeXLinkBox{\phantom{\rule{\n1}{\n2}}}}}
        }
    }
}

\usepackage{pgfplots}
\usepackage{anyfontsize}
\usepgfplotslibrary{groupplots}
\pgfplotsset{compat=1.18}
\definecolor{darkgreen06827}{RGB}{0,68,27}
\definecolor{darkgrey176}{RGB}{176,176,176}
\definecolor{grey}{RGB}{128,128,128}
\definecolor{lightgrey204}{RGB}{204,204,204}

\definecolor{darkblue}{RGB}{0,0,139}
\definecolor{darkgrey176}{RGB}{176,176,176}
\definecolor{darkred}{RGB}{139,0,0}
\definecolor{firebrick}{RGB}{178,34,34}
\definecolor{grey}{RGB}{128,128,128}
\definecolor{lightgrey204}{RGB}{204,204,204}
\definecolor{navy}{RGB}{0,0,128}
\definecolor{darkgrey176}{RGB}{176,176,176}
\definecolor{lightgrey204}{RGB}{204,204,204}
\definecolor{orange}{RGB}{255,165,0}
\definecolor{steelblue31119180}{RGB}{31,119,180}

\def\mytikzscale{0.3}

\def\myglobalscale{0.75}
\pgfplotsset{
    every axis/.append style={font=\sffamily\fontsize{\dimexpr\fpeval{30*\myglobalscale}pt\relax}{\dimexpr\fpeval{30*1.2*\myglobalscale}pt\relax}\selectfont,},
    every axis x label/.append style={
       yshift=\fpeval{-0.25*\myglobalscale}cm
    },
        every axis y label/.append style={
    yshift=\fpeval{0*\myglobalscale}cm,align=center,text width=\fpeval{8*\myglobalscale}cm
    }
}

\pgfplotsset{
    subjLeg/.style={
        legend image code/.code={%
 \draw [mark repeat=2,mark phase=2,semithick, darkgreen06827]
plot coordinates {
(0cm,0cm)
(0.3cm,0cm)
(0.6cm,0cm)
};          \draw [mark repeat=2,mark phase=2,semithick, darkgreen06827]
plot coordinates {
(0cm,-0.1cm)
(0.3cm,-0.1cm)
(0.6cm,-0.1cm)
};
\draw [mark repeat=2,mark phase=2,semithick,darkgreen06827 ]
plot coordinates {
(0cm,0.1cm)
(0.3cm,0.1cm)
(0.6cm,0.1cm)
};
}
    },
}

\pgfplotsset{
    leftLeg/.style={
        legend image code/.code={%
 \draw [mark repeat=2,mark phase=2,semithick, red]
plot coordinates {
(0cm,0cm)
(0.3cm,0cm)
(0.6cm,0cm)
};          \draw [mark repeat=2,mark phase=2,semithick, firebrick, dash pattern=on 9.6pt off 2.4pt on 1.5pt off 2.4pt]
plot coordinates {
(0cm,-0.1cm)
(0.3cm,-0.1cm)
(0.6cm,-0.1cm)
};
\draw [mark repeat=2,mark phase=2,semithick,darkred, dash pattern=on 9.6pt off 2.4pt on 1.5pt off 2.4pt]
plot coordinates {
(0cm,0.1cm)
(0.3cm,0.1cm)
(0.6cm,0.1cm)
};
}
    },
}

\pgfplotsset{
    rightLeg/.style={
        legend image code/.code={%
 \draw [mark repeat=2,mark phase=2,semithick, blue]
plot coordinates {
(0cm,0cm)
(0.3cm,0cm)
(0.6cm,0cm)
};          \draw [mark repeat=2,mark phase=2,semithick, navy, dash pattern=on 9.6pt off 2.4pt on 1.5pt off 2.4pt]
plot coordinates {
(0cm,-0.1cm)
(0.3cm,-0.1cm)
(0.6cm,-0.1cm)
};
\draw [mark repeat=2,mark phase=2,semithick,darkblue, dash pattern=on 9.6pt off 2.4pt on 1.5pt off 2.4pt]
plot coordinates {
(0cm,0.1cm)
(0.3cm,0.1cm)
(0.6cm,0.1cm)
};
}
    },
}

\pgfplotsset{
aread/.style={
legend image code/.code={
\draw [#1, \refcolor, fill=\refcolor, opacity=0.9] (0cm,-0.1cm) rectangle (0.6cm,0.1cm);
},
}
}

\newcommand{\refrefref}[2]{\LARGE {\begin{tabularx}{5.5cm}{Xr}#1&#2\\\end{tabularx}}}

\def\leftText{\refrefref{Left}{Mean $\pm$ 1 sd }}
\def\rightText{\refrefref{Right}{Mean $\pm$ 1 sd}}
\def\referenceText{\refrefref{Reference}{$\pm$ 1 sd}}
\def\referenceSubsText{\refrefref{Subjects}{1-6}}

\pgfplotsset{
every axis legend/.append style={
    cells={anchor=west},
},
}

\def\squatCycleName{Squat Cycle}

\usepackage{acro}

\DeclareAcronym{IK}{
  short = IK ,
  long  = Inverse Kinematics
}

\DeclareAcronym{ID}{
  short = ID ,
  long  = Inverse Dynamics
}

\DeclareAcronym{SO}{
  short = SO ,
  long  = Static Optimization
}
\DeclareAcronym{IMU}{
  short = IMU ,
  long  = Inertial Measurement Unit
}

\DeclareAcronym{GRF}{
  short = GRF ,
  long  = Ground Reaction Force
}

\DeclareAcronym{URDF}{
  short = URDF ,
  long  = Unified Robot Description Format ,
  cite = ioan_sucan_urdf_2009
}

\DeclareAcronym{opensimrt}{
  short = OpenSimRT ,
  long  = OpenSimRT ,
  cite = stanev_real-time_2021
}

\DeclareAcronym{ROS}{
  short = ROS ,
  long  = Robotics Operating System
}

\DeclareAcronym{VR}{
  short = VR ,
  long  = Virtual Reality
}

\DeclareAcronym{AR}{
  short = AR ,
  long  = Augmented Reality
}

\DeclareAcronym{COP}{
  short = COP ,
  long  = Center of Pressure
}

\DeclareAcronym{CFT}{
  short = CFT ,
  long  = Coordinate frame transformation
}

\DeclareAcronym{TF}{
  short = TF ,
  long  = tf: The transform Library,
  cite = tf_ros_2013_6556373
}

\DeclareAcronym{mocap}{
  short = Mocap ,
  long  = Motion Capture
}

\DeclareAcronym{STS}{
  short = STS ,
  long  = Sit-to-stand Stand-to-sit
}

\DeclareAcronym{EMG}{
  short = EMG ,
  long  = Electromyography
}

\DeclareAcronym{RMSE}{
  short = RMSE ,
  long  = Root Mean Square Error
}

\DeclareAcronym{sd}{
  short = sd ,
  long  = standard deviation
}

\DeclareAcronym{cgm}{
  short = CGM2.3 ,
  long  = Conventional Gait Model 2.3,
  cite = cgm2.3
}

\def \refcolor{lightgray}

\usepgfplotslibrary{fillbetween}

\newcommand{\GRAPHREFx}[2]{
 \addplot[name path=#1#2xrefmeanminus,draw=none] table [x=t, y=mmsx] {ref3dNi/#1#2.dat};  \addplot  [name path=#1#2xrefmeanplus,draw=none] table [x=t, y=mpsx] {ref3dNi/#1#2.dat};
 \tikzfillbetween [of= #1#2xrefmeanplus and #1#2xrefmeanminus, on layer=main] {\refcolor};}

\newcommand{\GRAPHREFy}[2]{
 \addplot[name path=#1#2yrefmeanminus,draw=none] table [x=t, y=mmsy] {ref3dNi/#1#2.dat};  \addplot  [name path=#1#2yrefmeanplus,draw=none] table [x=t, y=mpsy] {ref3dNi/#1#2.dat};
 \tikzfillbetween [of= #1#2yrefmeanplus and #1#2yrefmeanminus, on layer=main] {\refcolor};}

\newcommand{\GRAPHREFz}[2]{
 \addplot[name path=#1#2zrefmeanminus,draw=none] table [x=t, y=mmsz] {ref3dNi/#1#2.dat};  \addplot  [name path=#1#2zrefmeanplus,draw=none] table [x=t, y=mpsz] {ref3dNi/#1#2.dat};
 \tikzfillbetween [of= #1#2zrefmeanplus and #1#2zrefmeanminus, on layer=main] {\refcolor};}

\def \xxxd{
xmin=-0.05, xmax=1.05,
xtick style={color=black},
xtick={0,0.25,0.5,0.75,1},
xticklabels={0,25,50,75,100}
}

\begin{document}
\title{A real-time full-chain wearable sensor-based musculoskeletal simulation: an OpenSim-ROS Integration}
\author{Frederico Belmonte Klein, Zhaoyuan Wan, Huawei Wang, and Ruoli Wang
\thanks{This work was supported in part by the Swedish Research Council under Grant 2022-03268, Digital Futures Research Pair and WASP-WISE joint project (corresponding author: Ruoli Wang).}
\thanks{Frederico Belmonte Klein, Zhaoyuan Wan and Ruoli Wang are with KTH MoveAbility, Department of Engineering Mechanics, Royal Institute of Technology, SE-100 44 Stockholm Sweden (e-mail: frekle@kth.se; ruoli@kth.se).}
\thanks{Huawei Wang, is with the Department of Biomechanical Engineering, University of Twente, Enschede, The Netherlands and with wearM.AI BV., the Netherlands (e-mail: huawei.wang.buaa@gmail.com).}
}

\maketitle

\begin{abstract}

Musculoskeletal modeling and simulations enable the accurate description and analysis of the movement of biological systems with applications such as rehabilitation assessment, prosthesis, and exoskeleton design. However, the widespread usage of these techniques is limited by costly sensors, laboratory-based setups, computationally demanding processes, and the use of diverse software tools that often lack seamless integration. In this work, we address these limitations by proposing an integrated, real-time framework for musculoskeletal modeling and simulations that leverages OpenSimRT, the robotics operating system (ROS), and wearable sensors. As a proof-of-concept, we demonstrate that this framework can reasonably well describe inverse kinematics of both lower and upper body using either inertial measurement units or fiducial markers. Additionally, we show that it can effectively estimate inverse dynamics of the ankle joint and muscle activations of major lower limb muscles during daily activities, including walking, squatting and sit to stand, stand to sit when combined with pressure insoles. We believe this work lays the groundwork for further studies with more complex real-time and wearable sensor-based human movement analysis systems and holds potential to advance technologies in rehabilitation, robotics and exoskeleton designs.
\end{abstract}

\section{Introduction}
\label{sec:introduction}

\IEEEPARstart{A}{ccurate} description of human movement includes a comprehensive analysis of different components of the human body involved in performing physical actions, such as body postures, joint kinematics and kinetics, and muscle forces. Such analysis is not only fundamental for understanding the biomechanics of movement but also critical for enabling a wide range of applications. However, several challenges remain. A comprehensive movement analysis is typically performed in specialized laboratories and limited to a small number of accessible participants. Moreover, many internal biomechanical physical quantities are not always measurable in vivo. 
To address these limitations, musculoskeletal models and simulation frameworks (such as AnyBody~\cite{rasmussen_anybody_2003} and Open\-Sim~\cite{delp_opensim_2007,seth_opensim_2018}) are thus often adopted to estimate these unmeasurable quantities and to improve our understanding of the musculoskeletal system. These tools have been applied in diverse fields 
~\cite{domalain_was_2017}, 
~\cite{krakauer_motor_2006}, 
~\cite{bregman_effect_2011} 
and to facilitate the development of novel rehabilitation interventions and technologies. Nevertheless, due to the inherent complexity of biological systems and computational demands, it is a common practice that simulation procedures are carried out offline and based on dedicated laboratory settings. This presents two major obstacles for real-time clinical assessment and biofeedback application.

To bridge the gap and take advantage of the open-source nature of OpenSim, several frameworks have been proposed to enable real-time operation. Notably, RTOsim~\cite{pizzolato_real-time_2017} and OpenSenseRT~\cite{slade_open-source_2021} were developed to provide real-time estimations of \ac{IK} and \ac{ID} using \ac{mocap} system and \ac{IMU} input, respectively. Building on these efforts, OpenSimRT~\cite{stanev_real-time_2021} extended the computational pipeline of RTOsim, by incorporating \ac{IMU} input data from OpenSenseRT, while adding \ac{GRF}s and moments (GRFM) prediction as well as a fast implementation of muscle activation estimations using \ac{SO}. However, these frameworks do not address potential integration issues, such as interfacing with additional sensors (e.g., as multiple different \ac{IMU} drivers) or the implementation of other concurrent algorithms (such as visual feedback using  \ac{AR}). 

Among wearable sensors, \ac{IMU} have been extensively explored for real-time movement analysis~\cite{prasanth_wearable_2021}. However, when relying solely on \ac{IMU}s, many of these studies are either unable to implement the complete biomechanical analysis pipeline including \ac{IK}, \ac{ID} and \ac{SO} ~\cite{van_den_bogert_real-time_2013,lugris_human_2023} or they fail to achieve real-time performance due to the challenge in integrating data from multiple types of sensors~\cite{wang_wearable_2023}. To be used in an out-of-lab environment, such as clinics or for real-time activity monitoring, a fully wearable system incorporating wearable sensors such as \ac{IMU}s and pressure insoles, enabling real-time data streaming, is highly desired.  

Apart from \ac{IMU}s, \ac{AR} markers used as visual fiducials~\cite{pasman_implementation_2003} can serve as a highly cost-effective alternative for motion tracking. These markers require only a standard RGB camera and can provide both translational and rotational information of the marker with respect to the camera. \ac{AR} markers have been used in various applications, including orthopedic surgery, aiming to enhance the performance of surgeons~\cite{furman_augmented_2021,leon-munoz_integration_2023}. An implementation from~\cite{nagymate_affordable_2019} using \ac{AR} markers to compute lower body joint kinematics, reported angle errors of less than 7 degrees compared to a \ac{mocap} system during walking, an error magnitude comparable to soft tissue artifacts inherent to a \ac{mocap}-based measurement~\cite{leardini_human_2005}.

To achieve multiple-sensor integration and real-time operation, the structure of an ideal real-time pipeline aligns well with already implemented middleware used in robotics frameworks such as \ac{ROS}~\cite{quigley_ros_2009}, which serve as information brokers to facilitate seamless interfacing between different systems~\cite{tsardoulias_robotic_2017}. In the context of a wearable sensor-based musculoskeletal simulation, \ac{ROS} supports code reuse, simplifies the integration of sensors, enable simulation workflow, and facilitates system extension. In addition, its ability to support the development of reusable software components for hardware abstraction, inter-process communication, and data visualization makes it applicable across a wide range of applications. However, the absence of biological motion-specific features in \ac{ROS} has limited its adoption in biological systems. This limitation can be addressed by integrating OpenSim, which provides physiologically validated joint and muscle models specifically tailored for biomechanical analysis. To the best of our knowledge, no such integrated system exists.

The purpose of this study is therefore to propose a novel proof-of-concept framework that integrates OpenSim with \ac{ROS}, providing a platform for rapid prototyping of complex, real-time human movement analysis systems. The proposed wearable framework supports full-chain biomechanical analysis, including \ac{IK}, \ac{ID}, and muscle activation estimation. While similar platform for real-time biomechanical analysis of movement might exist, to the best of our knowledge, none offer full wearability or ROS integrated. A preliminary evaluation of this platform was conducted with healthy participants performing a range of movements, including arm and trunk motion, gait, sit-to-stand, stand-to-sit, and squats. The accuracy of estimated biomechanical parameters was assessed by comparison with reference data obtained simultaneously from a \ac{mocap} system. The proposed framework demonstrates strong potential as a flexible and extensible tool for future applications such in rehabilitation, assistive device development, and human-robot interaction.

\section{Materials and Methods}


\subsection{System overview}\label{sec:sys_overview}

The internal structure and interconnection of the different system components are shown in Fig.~\ref{fig:over}. The core pipeline consists of four main units, namely: (1) \ac{IK} (with \ac{IMU} or \ac{AR} markers); (2) external forces (with pressure insoles); (3) \ac{ID}, and (4) multithread \ac{SO}.

Here we primarily focus on modifications to the \ac{opensimrt} framework to make it modular, fully wearable, temporally and spatially aligned and compatible with \ac{ROS}. 
Given that the main challenge lies in integrating multiple different systems, we deployed the platform in a containerized format using Docker~\cite{merkel2014docker} to facilitate its usage and reproducibility. The container includes both \ac{ROS} and \ac{opensimrt}, with our system  encapsulates \ac{opensimrt} functionalities as \ac{ROS} nodes.
\footnote{All tests were performed using an Intel(R) Core(TM) i7-10750H CPU running Debian GNU/Linux 11 (bullseye).}

\begin{figure*}
\centering
\resizebox{\textwidth}{!} {
\input{tikz/system}
}
\caption{Block diagram of the system showing OpenSimRT-\ac{ROS} structure. The sensor inputs can be seen on the left. The top-left and bottom show sensors (either \acf{IMU}s or Video, for \acf{AR} marker input) for kinematics, and the pressure sensor, respectively. \acf{TF} blocks represent either reading or writing one or multiple time-stamped \ac{CFT}. The bold line symbolizes the main pipeline pathway from \acf{IK}, through \acf{ID} to \acf{SO}. On the right is the system visualization output. In light-green model-specific files and their creation order: from .osim model, create the moment-arm library; create an \ac{URDF} model and from the .osim and \ac{URDF} model create the joint mapping.}
\label{fig:over}\end{figure*}

\subsection{Musculoskeletal models}\label{subsec:offlinetasks}\hypertarget{subsec:offlinetasks}{}

To operate within the \ac{ROS} environment, both a \ac{URDF} model and a symbolic moment-arm library must be generated based on an OpenSim model for each subject. In this study, a generic model, gait2392~\cite{john_contributions_2012} was scaled to match the anthropometry of the subject~\cite{delp_opensim_2007} using marker data acquired by the \ac{mocap} system. The \ac{URDF} model was generated from a scaled gait2392 model using Pinocchio~\cite{LAAS_pinocchio_2022}. For the \ac{SO} node, a symbolic moment-arm library was generated based on the scaled OpenSim model, using an automated Python script provided by \ac{opensimrt}.

\subsection{System Components}
\subsubsection{Inverse Kinematics}\clabel{subsec:ik}

The \ac{IK} node estimates generalized joint angle from orientation input, constrained by a kinematic chain model described in the \ac{opensimrt}. To generalize orientation inputs beyond \ac{IMU} input from OpenSimRT, the \ac{ROS} \ac{TF} was used as a generic wrapper to support both \ac{AR} markers~\cite{scott_niekum_ar_track_alvar_2012} and \ac{IMU}.  The resulting model-constrained joint angles (${\bm{q}}$) were then spline-filtered~\cite{WOOD1979477} and published alongside with filtered joint velocities (${\bm{\dot{q}}}$) and accelerations (${\bm{\ddot{q}}}$). This node publishes time-dependent \ac{CFT} between adjacent body segments (i.e. between foot and tibia) via the \ac{TF} system using the \ac{ROS} and \ac{URDF} model. Conversion between the OpenSim and \ac{URDF} was achieved through a predefined joint mapping. The resulting \ac{CFT}s were subsequently used to calculate the application points for external forces.

For the \ac{IK}, an initial static calibration procedure is required to determine a fixed transformation between the sensor orientation and the corresponding body segment. This calibration was conducted using 10 frames of orientations, which were then averaged using a quaternion singular value decomposition-based algorithm~\cite{curtis_note_1993} to correct for heading offset in both the \ac{IMU}s or \ac{AR} markers.

\subsubsection{Insole driver}\clabel{subsec:insole}
The insole driver module serves as an interface between the pressure insole data and the biomechanical analysis framework. A python-based \ac{ROS} node, implemented using Moticon SDK (Endpoint SDK, Moticon ReGo AG, Germany), converts the estimated normal \ac{GRF} and \ac{COP} into \ac{ROS} messages. As the insole data is streamed in bursts of irregular length, the insole driver is therefore designed to include filtering and time correction mechanisms to account for time offset. It is also worth noting that the insole input may not be temporally aligned with the joint angle data within the \ac{ROS} system, necessitating an additional time synchronization step (see Sec.~\ref{sec:sync}).

\subsubsection{Inverse Dynamics node}\clabel{subsec:asyncid}

\ac{ID} node receives inputs from \ac{IK} and the \ac{GRF}s to calculate generalized joint torques (referred to as wrenches in \ac{ROS}) using the recursive Newton-Euler method. The \ac{COP} is originally expressed in a local coordinate frame of the insole, where the x-axis represents the anteroposterior direction (heel to toe), and the y-axis represents the mediolateral (side to side). For \ac{ID} calculations, \ac{COP} estimates from the insoles are converted with \ac{CFT}s and expressed in the global coordinate frame. The transformed \ac{GRF} (wrench) was applied as a force vector normal to the ground during movements. 


Buffers for both \ac{IK} and \ac{ID} wrenches, along with \ac{CFT}s from the appropriate time-stamp (see \ref{sec:sync} for more details) were used to achieve the proper temporal and spatial alignment between the \ac{IK} and external forces. This node then publishes synchronized \ac{IK} and \ac{ID} messages,  i.e. combined joint angles and joint torques - referred to as joint states in \ac{ROS} terminology.

\subsubsection{Multithreaded Static Optimization Node}\clabel{subsec:sorr}

This node addresses the muscle redundancy problem using \ac{SO}, minimizing the sum of squared muscle activations~\cite{stanev_real-time_2021}, which reflects metabolic energy consumption~\cite{thomas_k_uchida_biomechanics_2020}. The \ac{SO} is executed in parallel using a deterministic multithreaded pipeline scheduler. Each combined \ac{IK} and \ac{ID} message is sequentially assigned to one of N threads using a modulo operation based on the message count index. Each thread uses its own optimizer warm start values. For example, when operating at 100Hz with four threads, the initial guess for each thread was offset by three samples, resulting a slightly increased computation time per thread. A final synchronization deadline is defined for the time sequencer, which also serves visualization and graphing purposes. If the optimizer fails to converge before this deadline, the corresponding sample is discarded. 

\subsection{Synchronization}\label{sec:sync}

The insole SDK multiplexes data from the left and right insoles using a single socket connection, delivering bursts of variable length. Before usage, the incoming messages are sorted, re-stamped, and buffered. Timestamps for the left and right insole channel are assigned based on a previously established synchronization event and the internal time clocks of each insole sensor. Finally, spline filtering is applied to the normal forces and \ac{COP} data.

\subsection{Data Acquisition}\clabel{sec:data_aq}
\subsubsection{Calibration}

For the lower limb model, the calibration position was defined as standing upright with feet parallel and spaced at pelvic width. For the upper limb model, the calibration position was defined as standing upright with upper limbs parallel to the torso, palms facing inwards and parallel to the torso~\cite{yoshida_three-dimensional_2022}. The calibration step is necessary for establishing the correspondence between the orientation axes of the \acp{IMU} and those of the subject's joints.

\subsubsection{Experimental Setup}
The experimental setup included two stand-alone measurement systems: the laboratory-based \acp{mocap} system and the wearable sensor system. The \acp{mocap} system included 10 infrared cameras (VICON Vantage V16, Vicon, UK), four force plates (AMTI Optima, AMTI, USA) and six electromyography (EMG) sensors (Pico EMG, Cometa, Italy). Reflective marker trajectories, GRFs and muscle activations were recorded simultaneously at 100 Hz, 1000 Hz, and 1000 Hz, respectively. In \ref{sec:pipe:complete}, two systems were operated in parallel. 

For the wearable sensor system, a webcamera, ACR010 (Acer, Taiwan) was used as a video source (640x480 @30fps) for testing \ac{AR} markers as the input (\ref{sec:pipe:ar}). For evaluating the system with \ac{IMU} sensor as the input (\ref{sec:pipe:imu_ik_only}), XIMU-3 (x-io Technologies Limited, UK) was used, streaming data via User Datagram Protocol (UDP) over WiFi at 100Hz. For all \ac{IMU}s, orientation was provided by the sensor's internal algorithm based on data from integrated 3D accelerometers and 3D gyroscopes. The initial heading was set to zero at startup, and all \ac{IMU}s were aligned in a line, facing the same direction. Pressure insoles (Moticon OpenGo version 3, Moticon ReGo AG, Germany) with 16 pressure sensors, were used to provide real-time normal force data at 100Hz via a Bluetooth (\ref{sec:pipe:complete}). The insoles transmitted data to an Android device, which was connected to the main system via WiFi. Prior to measurement, the insoles were warmed up to body temperature and calibrated to the subject's weight as instructed. 


\subsection{Bench Testing of the Framework}\label{sec:pipe_testing}

The framework was tested with three different setups: 1) the upper body kinematic using \ac{AR} markers; 2) walking kinematic analysis using \ac{IMU}s; and 3) full-chain biomechanical analysis, including \ac{IK}, \ac{ID}, and muscle \ac{SO}, with simultaneously data acquisition from \ac{IMU}s and pressure insoles, and the Mocap system. An additional test was done on recorded data to evaluate processing times and potential delays in the full-chain configuration. All experimental sessions were conducted at the Promobilia MoveAbility Lab. The study was approved by the Swedish Ethical Review Authority (2020-02311).


\subsubsection{Upper body inverse kinematics using Augmented Reality Markers}\label{sec:pipe:ar}

\begin{figure}
\begin{subfigure}{0.3\columnwidth}
\includegraphics[height=3.5cm]{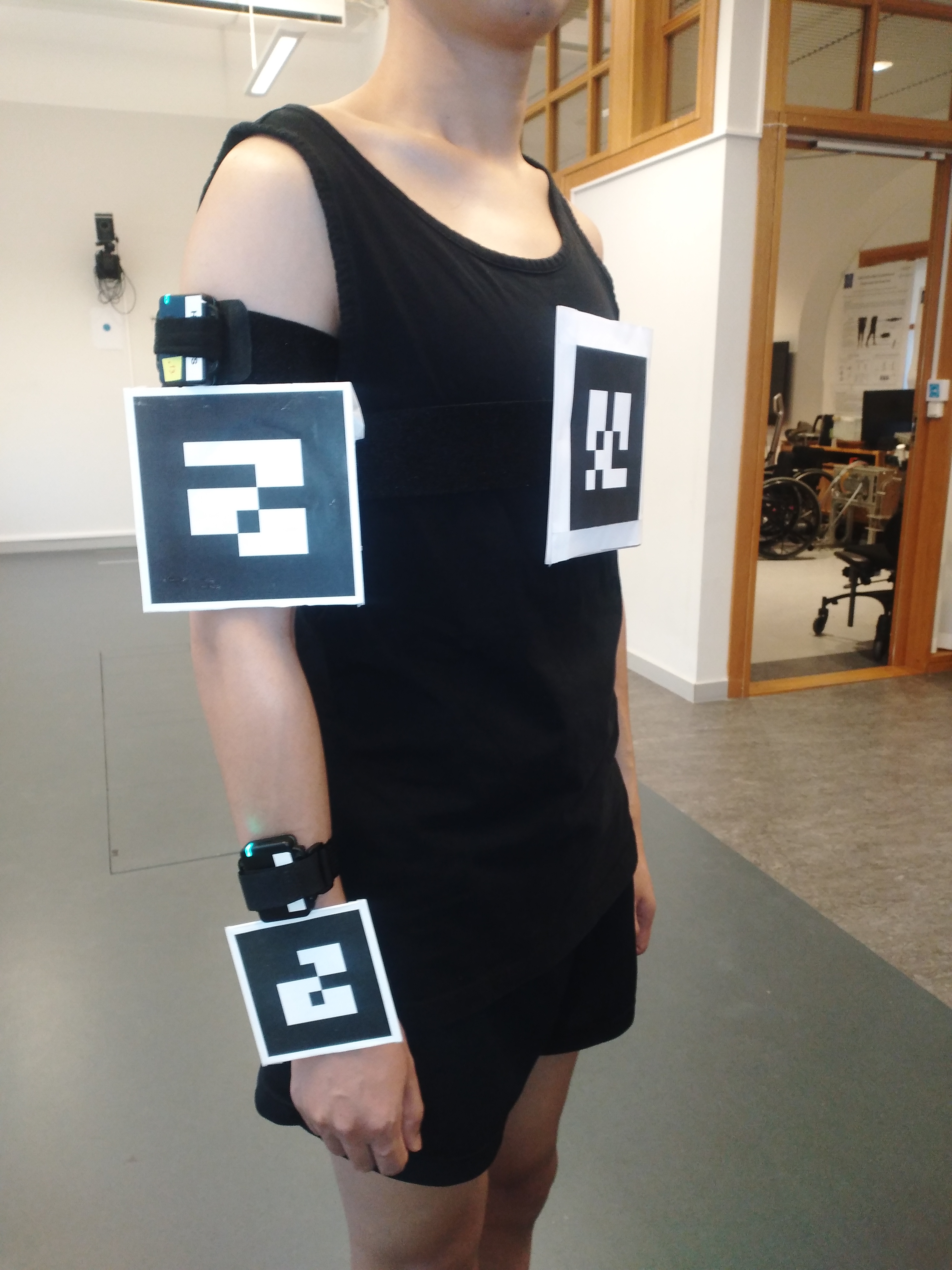}\caption{}\label{fig:ar:placement}
\end{subfigure}
\centering
\begin{subfigure}{0.65\columnwidth}
    \includegraphics[height=3.5cm]{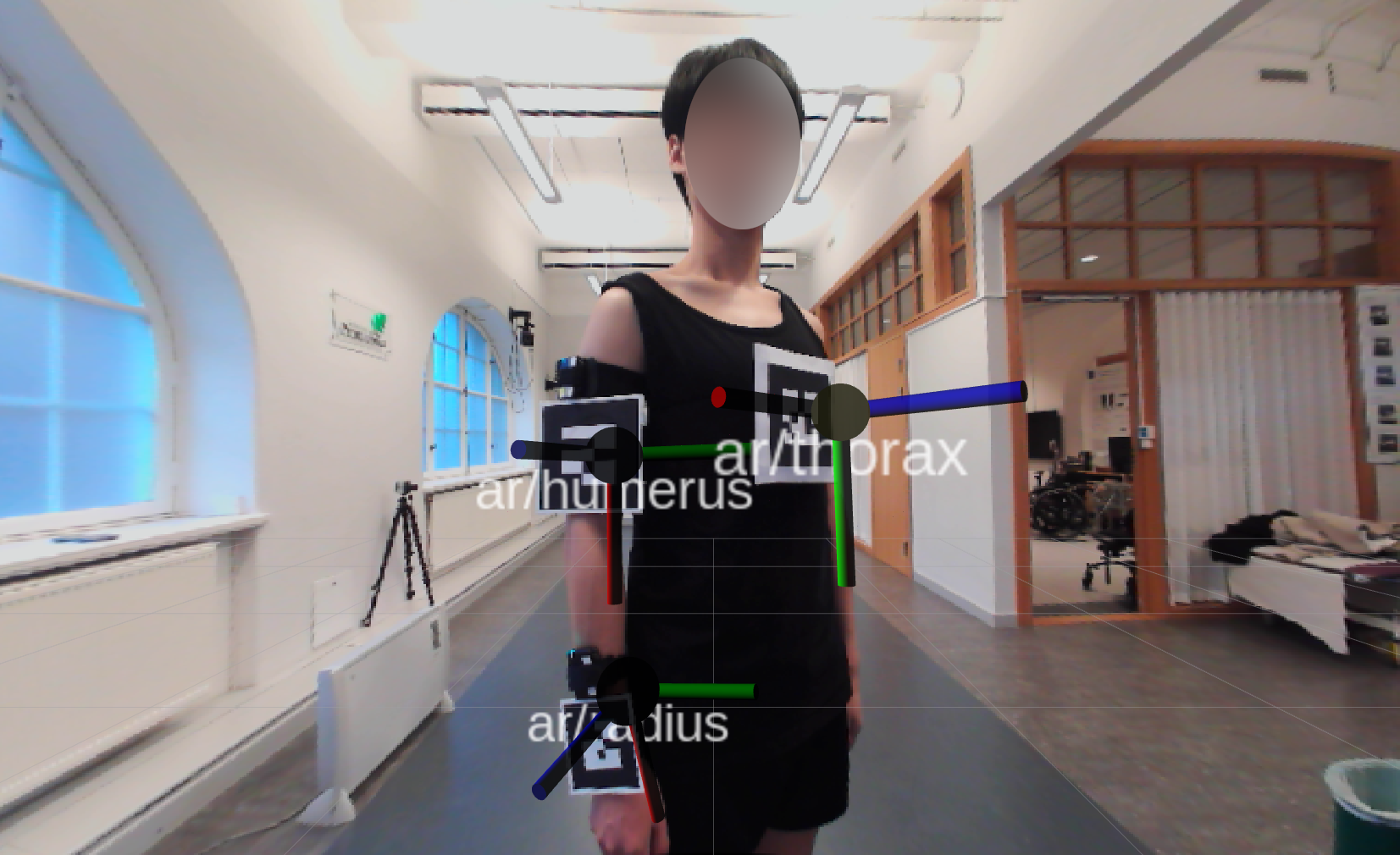}\caption{}\label{fig:ar:cameraview}
\end{subfigure}
\caption{(\subref{fig:ar:placement}) One testing subject wearing \ac{AR} marker plates together with \ac{IMU}s standing in the calibration position. (\subref{fig:ar:cameraview}) The camera view showing coordinate frames from detected \ac{AR} markers. }
\label{fig:ar}
\end{figure}

Three \ac{AR} markers were placed on the forearm (\qtyproduct{9 x9}{\centi\meter}), upper arm (\qtyproduct{12 x 12}{\centi\meter}), and torso (\qtyproduct{13 x 13}{\centi\meter}) and the video camera was set \qty[round-precision = 1]{1}{\meter} in front of the subject (Fig.~\ref{fig:ar}). The \ac{AR} markers were printed on standard office paper with an inkjet printer and taped to 3D-printed flat surface. For comparison, three \ac{IMU}s were placed adjacent to each corresponding \ac{AR} marker. The \ac{AR} marker position and sizes were selected to optimize visibility for camera view and to maximize tracking accuracy.

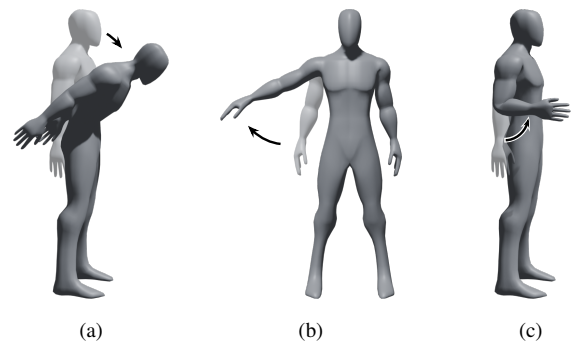
\begin{figure}
\centering
	\begin{subfigure}{0.33\columnwidth}
	  \centering
\resizebox{!}{4cm}
{
	    \begin{tikzpicture}

	    \node (a) {\includegraphics[height=6.cm]{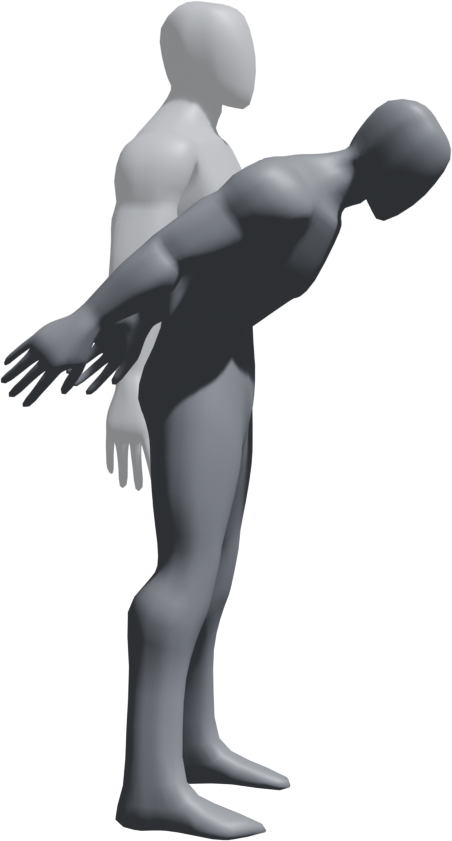}};

	    \path[draw=white,   line width=1mm, -{Stealth[length=10pt,bend]}]
		(0.3,2.5) arc  (60:30:1cm);

	    \path[draw=black, line width=0.5mm, -{Stealth[length=7pt,bend]}, shorten >=0.5mm, shorten <=0.3mm]
	      (0.3,2.5) arc  (60:30:1cm);

	    \end{tikzpicture}
    }
	    \caption{	  }\label{fig:ar_act:trunk}
	\end{subfigure}\begin{subfigure}{0.33\columnwidth}
  \centering

\resizebox{!}{4cm}
{
  \begin{tikzpicture}

    \node (a) {\includegraphics[height=6.cm]{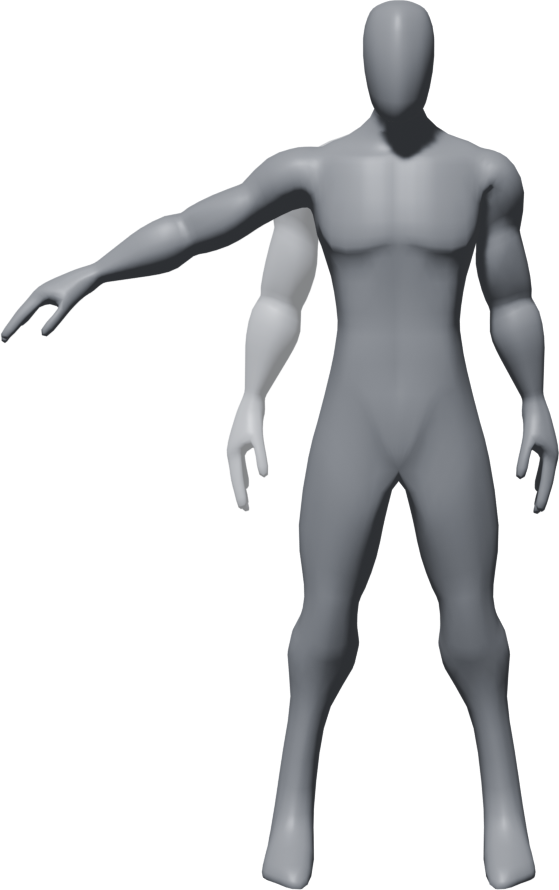}};

    \path[draw=white,   line width=1mm, -{Stealth[length=10pt,bend]}]
    (-0.6,0.2) arc  (265:215:1cm);

    \path[draw=black, line width=0.5mm, -{Stealth[length=7pt,bend]}, shorten >=0.5mm, shorten <=0.3mm]
     (-0.6,0.2) arc  (265:215:1cm);

\end{tikzpicture} }
  \caption{  }\label{fig:ar_act:elev}
\end{subfigure}\begin{subfigure}{0.33\columnwidth}
  \centering

\resizebox{!}{4cm}
{
  \begin{tikzpicture}

    \node (a) {\includegraphics[height=6.cm]{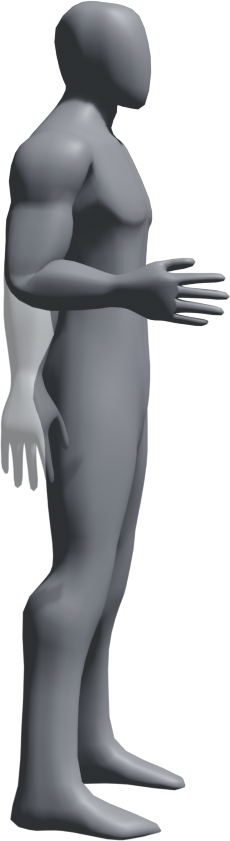}};

\path[draw=white,   line width=1mm, -{Stealth[length=8pt,bend]}]
(-0.55,0.28) arc  (270:360:0.5cm);

\path[draw=black, line width=0.5mm, -{Stealth[length=5pt,bend]}, shorten >=0.5mm, shorten <=0.3mm]
(-0.55,0.28) arc  (270:360:0.5cm);

    \end{tikzpicture}
}
    \caption{  }
  \label{fig:ar_act:biceps}
\end{subfigure}
\caption{Three single plane movements tested with \ac{AR} markers and upper body model:
trunk flexion/extension (\subref{fig:ar_act:trunk}),
lateral arm elevation (\subref{fig:ar_act:elev}) and
elbow flexion (\subref{fig:ar_act:biceps}).}
\label{fig:ar_act}
\end{figure}

The bench test was performed on a single subject (M, \age{27}, \height{1.75}, \weight{60}) using an unscaled upper body OpenSim model \texttt{mobl2016\_v03.osim} (modified from\cite{saul_benchmarking_2015}). The model consisted of three segments (thorax, humerus, and radius) and three joints (shoulder, elbow, and wrist). Following a calibration, the subject was instructed to perform three single-plane movements with 5 repetitions each (see fig.~\ref{fig:ar_act}). The movements included trunk flexion/extension (approx. \qty{30}{\degree}), lateral arm elevation (approx. \qty{45}{\degree}) and elbow flexion (approx. \qty{90}{\degree}). The \ac{RMSE} between \ac{AR}-based and \ac{IMU}-based joint angles was computed.

\subsubsection{Inverse Kinematics during walking with only IMUs}\label{sec:pipe:imu_ik_only}

\begin{figure}
\centering
\begin{subfigure}{0.3\columnwidth}
  \centering
  \includegraphics[height=4cm,trim={13.5cm 1cm 13.5cm 2cm},clip]{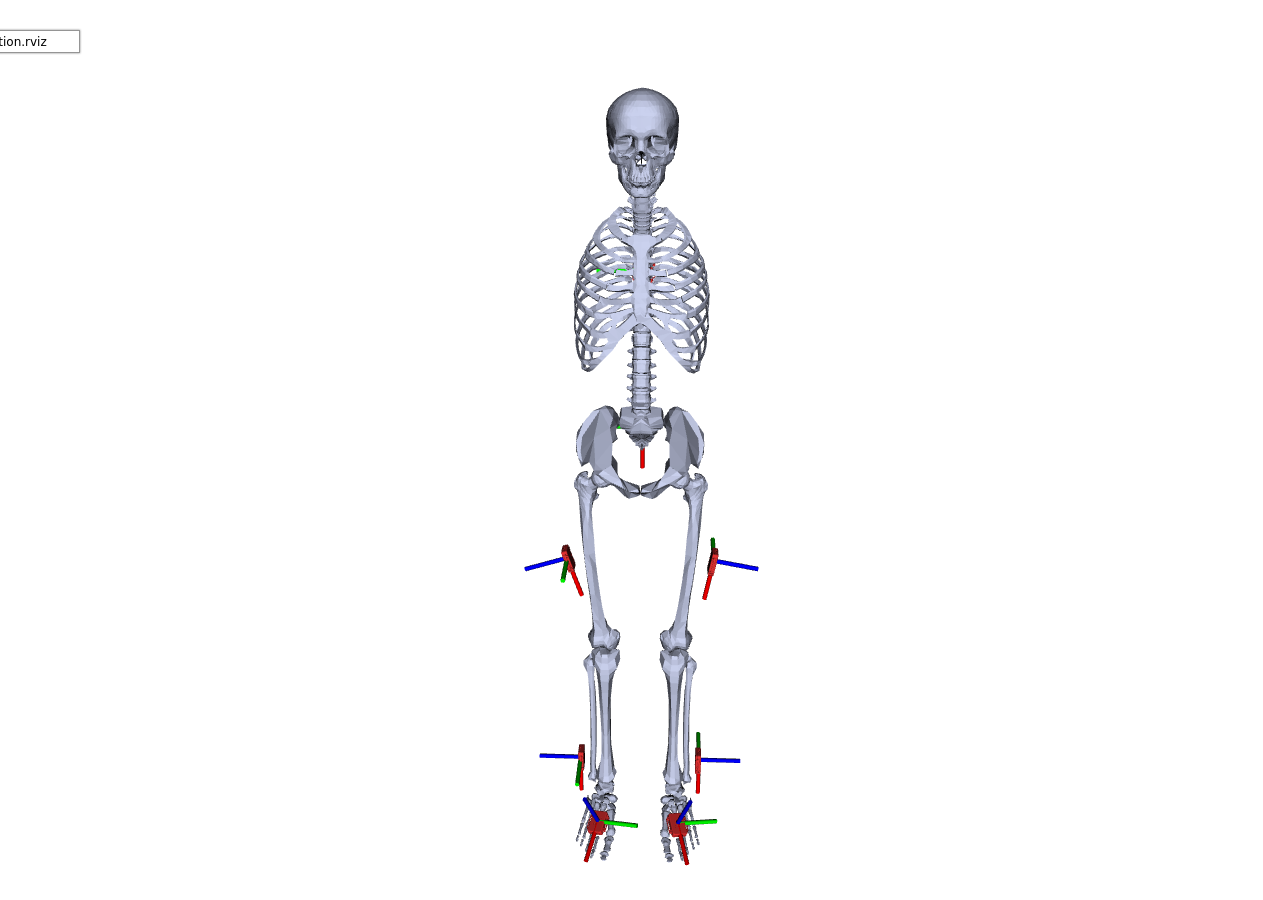}
  \caption{    }\label{fig:imu_ori:front}
\end{subfigure}\begin{subfigure}{0.3\columnwidth}
  \centering
  \includegraphics[height=4cm,trim={12cm 1cm 15cm 0cm},clip]{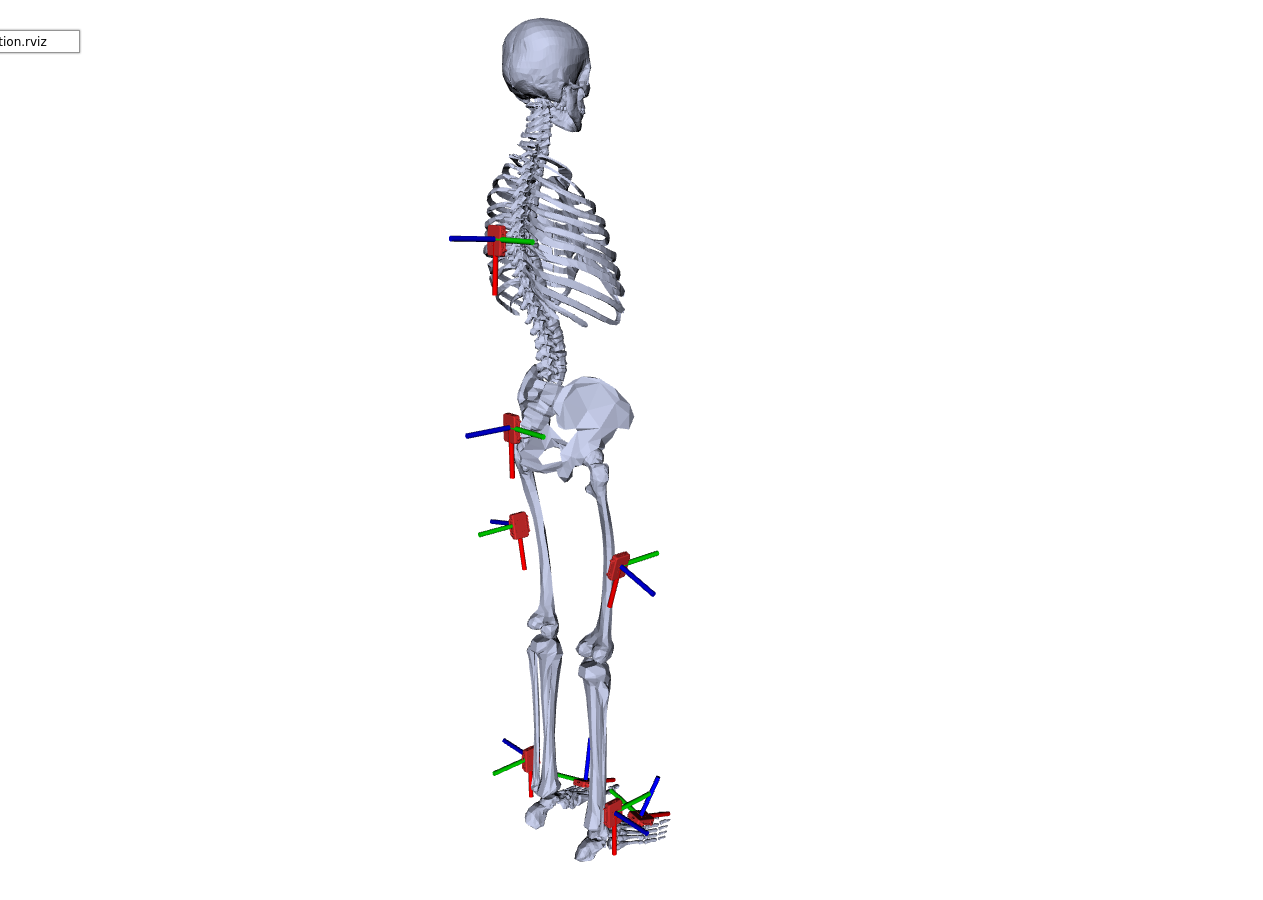}
  \caption{    }\label{fig:imu_ori:back}
\end{subfigure}\begin{subfigure}{0.3\columnwidth}
  \centering
  \includegraphics[width=0.6\linewidth,trim={9.5cm 8cm 10cm 0},clip]{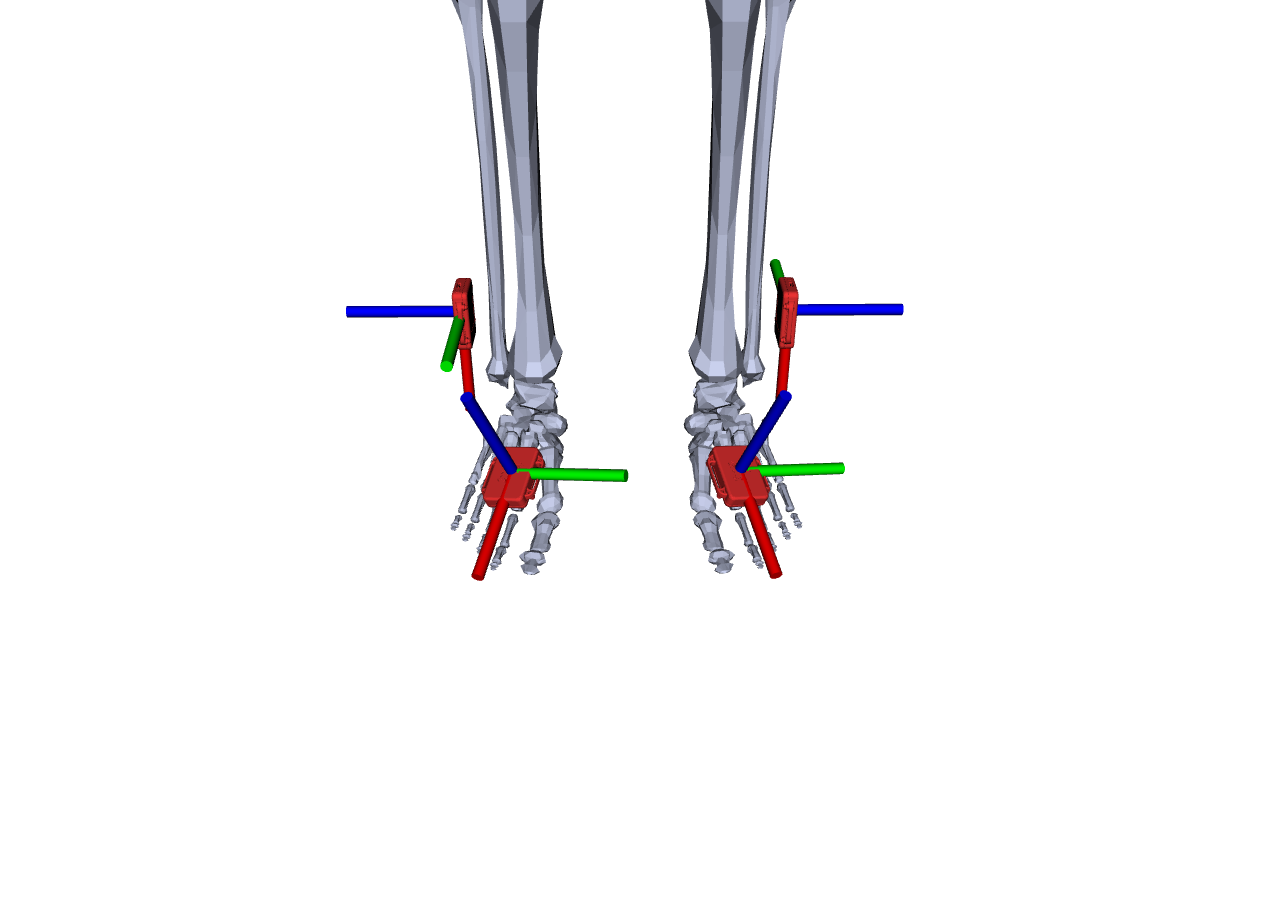}
  \caption{
    }
  \label{fig:imu_ori:feet}
\end{subfigure}

\caption{Illustration of \ac{IMU} placement (in red) on a full-body gait1992 \ac{URDF} model in front view~(\subref{fig:imu_ori:front}), back view~(\subref{fig:imu_ori:back}) and feet close-up view (\subref{fig:imu_ori:feet}). \ac{IMU} axis follows common color convention {\color{red}X}, {\color{green}Y}, {\color{blue}Z}. }
\label{fig:imu_ori}
\end{figure}

To evaluate the performance \ac{IMU}-based \ac{IK}, data were acquired from six able-bodied participants (5M/1F, \age{32.5 \pm 6.3}, \height{1.75 \pm0.09}, \weight{75 (12)}) using \ac{IMU} input only. Eight \ac{IMU} sensors were positioned as follows: posteriorly on torso, pelvis, laterally on each thigh and lower leg, and on top of each foot (see \cref{fig:imu_ori}) with custom-made IMU holders and flexible straps. During the calibration (see \cref{sec:data_aq}.1), subjects were instructed to remain as still as possible. Following calibration, each subject performed five trials of level ground walking at a self-selected speed in along a 9-meter walkway. Each trial consisted of 3 to 6 steps, which were manually segmented and normalized by 100\% gait cycle. Trials consisting few than three gait cycles were excluded from further analysis.  Joint kinematics were computed for: pelvis tilt, obliquity, and rotation; hip flexion/extension, abduction/adduction, and external/internal rotation; knee extension/flexion and ankle plantarflexion/dorsiflexion.

\subsubsection{Full-chain real-time biomechanical analysis with IMU and pressure insole}\label{sec:pipe:complete}

To evaluate the performance of full-chain pipeline using both \ac{IMU} and insole input, data was acquired from two able-bodied participants (1F, \age{42}, \height{1.63} and \weight{53} and 1M, \age{30}, \height{1.70} and \weight{72}). \ac{IMU}s data were acquired as described in \cref{sec:pipe:imu_ik_only}, concurrently with streamed data from the insole driver. The combined input simultaneously drove the \ac{IK}, \ac{ID} and \ac{SO} nodes in real-time. 

Participants wore shoes fitted with pressure insoles matched to their foot size. The insoles were calibrated to the subject's body weight. EMG sensors were placed on five major lower-limb muscles, i.e., tibialis anterior (TA), medial gastrocnemius (GM), soleus (SOL), gluteus maximus, and vastus medialis, following the SENIAM recommendations\cite{HERMENS2000361}. The skin was shaved and cleaned prior to placing the EMG electrodes. Reflective markers were placed on the trunk and lower body following the \ac{cgm} marker set (2.3 version of the Conventional Gait Model), with four additional markers on each shoe. 

The subjects were instructed to perform three common daily activities (see \cref{fig:pipe_full}): level ground walking at a self-selected speed, \ac{STS}, and squat. For level ground walking, automatic step segmentation was performed using a low pass filter applied to the insole force data, with a threshold set at \qty{10}{\percent} of the subject's body weight. All other trials were manually segmented.

\begin{itemize}
    \item Level ground walking: five trials 
    \item \ac{STS}: two trials with five repetitions
    \item Squat: two trials with five repetitions of moderate-depth squats
\end{itemize}


To synchronize \ac{CFT}s derived from \ac{IK} and pressure insole data, a buffer delay of \qty{260}{\milli\second} was added to \ac{IK} to account for the latencies associated with the insole data and the \ac{CFT} broadcaster  
\def\W{5.2}     
\def\L{1.6}     
\def\T{0.08}    
\def\H{2.2}     

\begin{figure}
    \centering
    \begin{subfigure}{0.32\columnwidth}
        \centering

        \walkingHuman
        \caption{}
        \label{fig:pipe_full:walk}
    \end{subfigure}%
    \begin{subfigure}{0.32\columnwidth}
        \centering
        \stsHuman
        \caption{}
        \label{fig:pipe_full:sts}
    \end{subfigure}%
    \begin{subfigure}{0.32\columnwidth}
        \centering
        \squatHuman
        \caption{}
        \label{fig:pipe_full:squat}
    \end{subfigure}
    \caption{Three activities executed with full-chain pipeline and lower body model:
walk (\subref{fig:pipe_full:walk}),
\acf{STS} (\subref{fig:pipe_full:sts}) and
squat (\subref{fig:pipe_full:squat}).}\label{fig:pipe_full}
\end{figure}

Recordings from the Mocap system were processed with the Vicon Nexus software. 
For squat, each repetition was defined as the interval between the onset of knee flexion and the return to full knee extension. 
For the \ac{STS}, each repetition began when the participant fully lifted off the stool and ended just before the body made contact with the stool.

Marker trajectories and ground reaction forces were used in OpenSim for an offline computation of \ac{IK} and \ac{ID}. Joint torque was normalized to the body weight. Raw \ac{EMG} recording were high-pass filtered at 20Hz, low-pass filtered at 450Hz, and a notch filtered at 50Hz, then rectified and smoothed using a 5 Hz low-pass filter. The resulting enveloped \ac{EMG} signals were normalized to the maximum value across all trials for the same participant, and down-sampled to 100 Hz to align with motion data. \ac{RMSE} was computed between outputs from our real-time pipeline and those from the offline OpenSim IK and ID analysis based on \ac{mocap} data. 


\subsubsection{Measuring the transport delays caused by ROS}\label{subsec:delay}
Measuring computational times and latency for the system's components is essential to demonstrate real-time performance within strict time constraints.
To evaluate transport delays and processing times within the \ac{ROS} system, we used the \texttt{ros::Time} class for precise measurements and extended our common message type to include a list of events timestamps to track key events along the main pipeline's critical path (bold line in \cref{fig:over} was tracked) .

For this test, \ac{IK} and insole data from a single subject (5 trials) during walking were played back the same sensor data recordings through an otherwise full pipeline including \ac{IK}, \ac{ID}, and \ac{SO}. As the \ac{SO} node is among the most computationally demanding components, requiring a multithreaded pipeline scheduler, we evaluated its processing times under different thread counts (N = 4, 6, or 12). The full list of tracked events (see below \cref{tab:met:timings}) was recorded from the combined output of \ac{SO} node in a \texttt{rosbag} before message reordering.
\begin{table}[H]
\caption{Event timing definitions\label{tab:met:timings}}
\begin{minipage}{\columnwidth}
\begin{tabularx}{\textwidth}{lcl}
\toprule
No.\parnote{Events 2 and 4 are filtering of IK and GRFMs respectively if those weren't already filtered signals. Here omitted, due to data already previously spline filtered.} & Node & Description \\
\midrule
0 & \ac{IK} & \variableEventA \\
1 & \ac{ID} & \variableEventB \\
3 & \ac{ID} & \variableEventD \\
5 & \ac{ID} & \variableEventF \\
6 & \ac{ID} & \variableEventG \\
7 & \ac{SO} & \variableEventH \\
8 & \ac{SO}[thread i] & \variableEventI \\
9 & \ac{SO}[thread i] & \variableEventJ \\
\bottomrule
\end{tabularx}
\parnotes
\end{minipage}
\end{table}

We evaluated the distribution of latencies between consecutive events and calculated mean event durations. Due to variability introduced by operating system scheduling, network transmission, and the nature of optimizers, the execution times vary.  To evaluate the worst-case scenarios, we 
also computed the 95th percentile latency,
representing the minimum time window within which \qty{95}{\percent} of messages were expected to be received for the critical path of the pipeline. This was done by computing the empirical cumulative distribution -- industry standard practice for latency monitoring in distributed systems\cite{zhang_performal_2023,chen_empirical_2017,eaton_distributed_2022} -- obtained from histograms from 1000 bins.

\section{Results}


\subsection{Upper body Inverse Kinematics}

\begin{figure}
\centering
  \resizebox{1\columnwidth}{!}{
\input{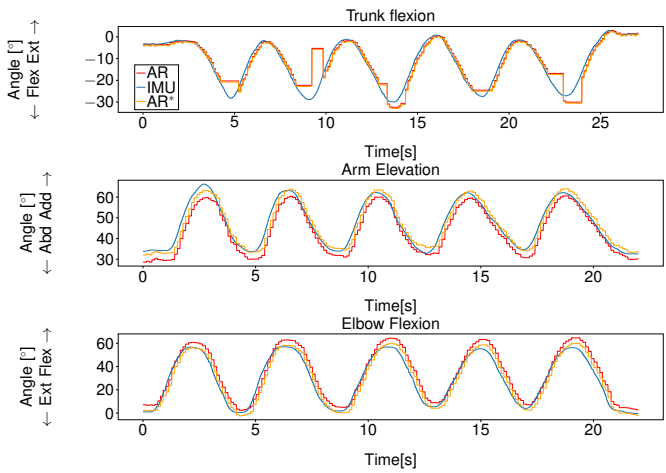}
}

\caption{Five repetitions of \artrunkmovementname~(top), right \arelevationmovementname~(middle) and right \arcurlmovementname~(bottom). The red line illustrates joint angles from \ac{AR} markers, blue line shows reference angle from \ac{IMU}s, and the yellow line (\ac{AR}$^{*}$) indicates joint angle from \ac{AR} markers after offset correction.}
\label{fig:res_ar}
\end{figure}

\cref{fig:res_ar} illustrates the real-time joint angle estimation of \artrunkmovementname, \arelevationmovementname~and \arcurlmovementname~with \ac{AR} markers and \ac{IMU}s. Overall, there is a good agreement between the \ac{IMU}-based and \ac{AR}-based joint angle measurements. An offset of approximately \ang{2} was observed in the \arelevationmovementname~ and \arcurlmovementname, but not in the \artrunkmovementname. After applying offset correction, compared to the \ac{IMU}-based angles, the RMSE was \ang{4.1} for trunk flexion and \ang{2.4} for \arelevationmovementname, and \ang{4.7} for \arcurlmovementname, respectively. It is worth noting that \ac{AR}-based estimates had a lower sampling frequency and occasional data lost, particularly during \artrunkmovementname~beyond \ang{20}.

\subsection{Walking Kinematics with IMUs}\label{sec:pipe:imu_ik_only:res}

\begin{figure}
\centering

\input{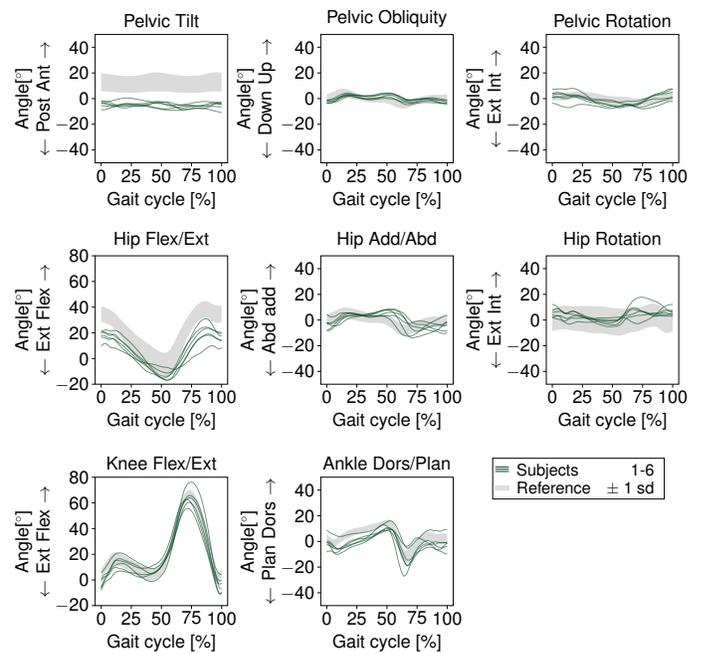}

\caption{\ac{IMU}-based real-time \ac{IK} during walking from six able-bodied subjects and an open-source reference data based on the Mocap system~\cite{reznick_lower-limb_2021}.
Solid lines represents the mean joint angles for each subject, averaged across both sides and five trials.}
\label{fig:ik_gait}
\end{figure}

\cref{fig:ik_gait} illustrates the real-time mean gait kinematics for both the left and right sides for six subjects. For comparison, reference curves were obtained from an open dataset based on the optical Mocap system~\cite{pinzone_comparison_2014}. Overall, the real-time \ac{IMU}-based joint kinematics showed good agreement with the reference data, particularly in the sagittal plane at the ankle and knee joints. However, a consistent offset of approximately \ang{15} was observed in both pelvic tilt and hip flexion/extension angles.

\subsection{Full-chain Real-time Lower Limb Pipeline with Pressure Insole and IMUs}

\begin{figure}[!ht]
\centering
{
\includegraphics[width=\columnwidth]{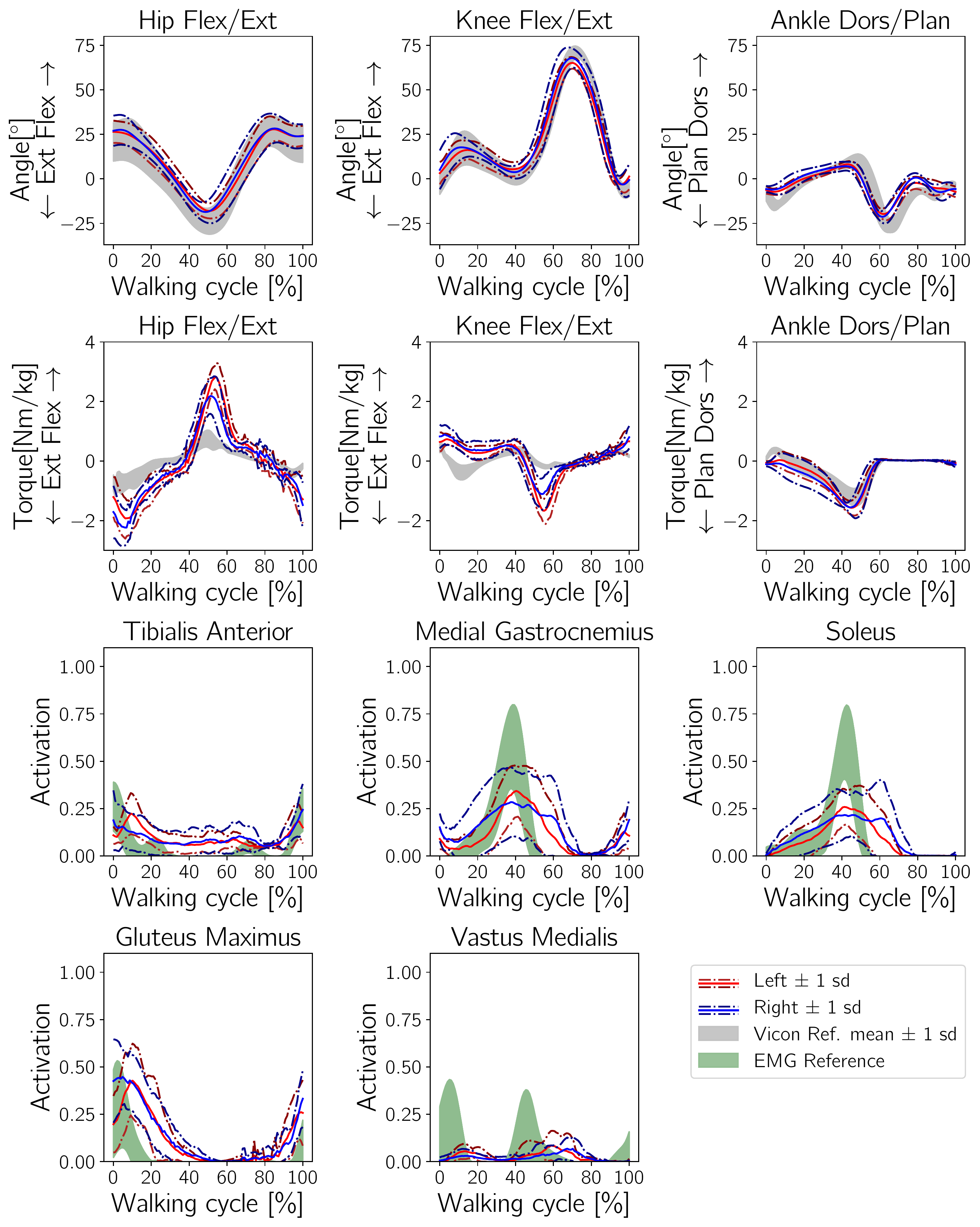}
}

\caption{Lower limb sagittal plane \ac{IK}, \ac{ID} at the ankle, knee and hip, along with muscle activation of tibialis anterior, medial gastrocnemius, soleus, gluteus maximus, and vastus medialis during level ground walking based on the full-chain real-time pipeline for two subjects. Solid and dashed lines illustrated mean $\pm 1$ \ac{sd} of the left and right sides. The data include 5 trials from each of the two subjects, totaling 10 left and 10 right gait cycles. Shaded areas indicate reference data obtained based on the motion capture system (gray) or \ac{EMG} (dark green).
}
\label{fig:res:walk}
\end{figure}

\begin{figure}[!ht]
\centering
 \includegraphics[width=\columnwidth]{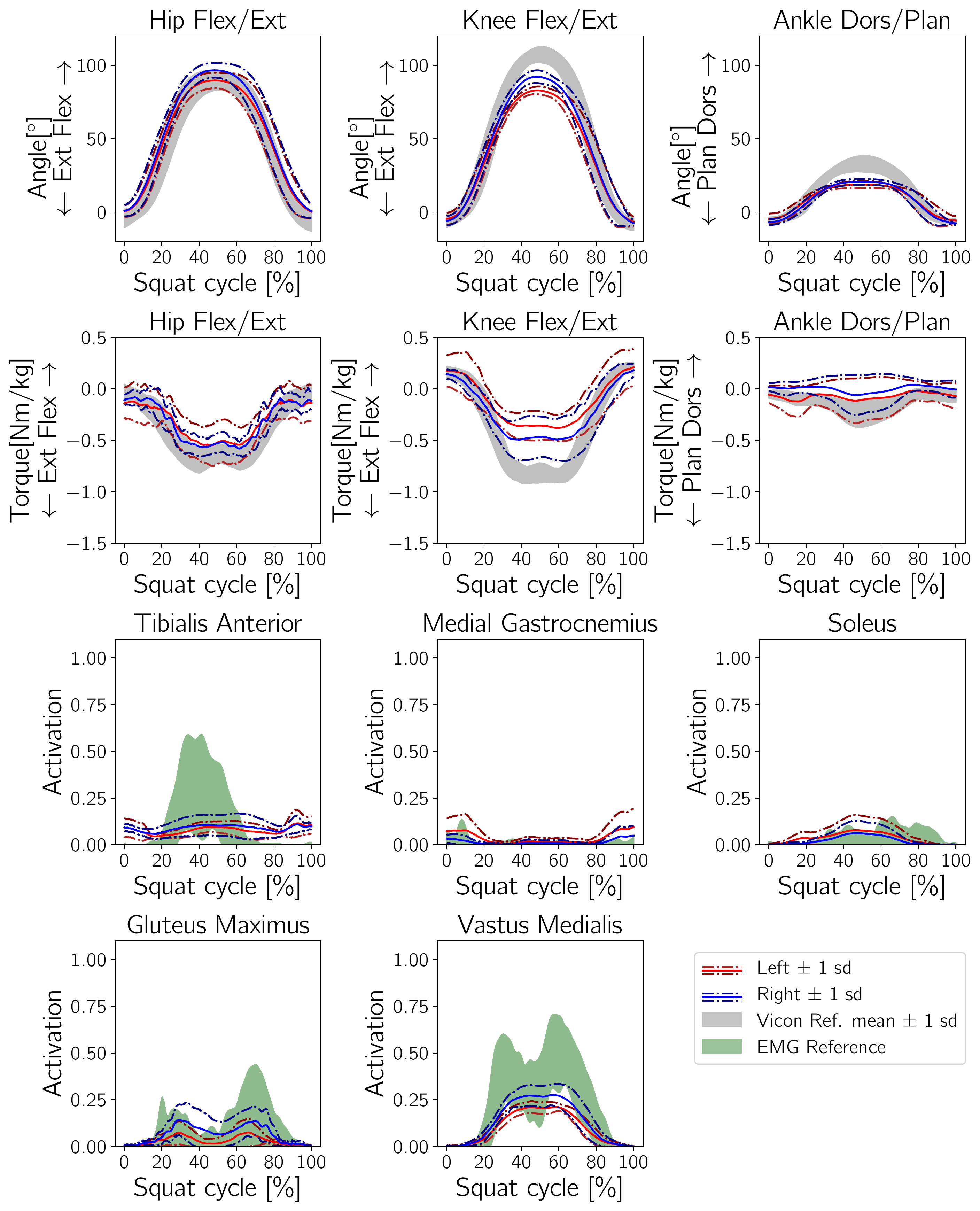}

\caption{Lower limb sagittal plane IK, ID at the ankle, knee and hip, along with muscle activations of tibialis anterior, medial gastrocnemius, soleus, gluteus maximus, and vastus medialis during squat based on the full-chain real-time pipeline for two subjects. Solid and dashed lines illustrated mean $\pm 1$ \ac{sd} of the left and right sides. Shaded areas indicate reference data obtained based on motion capture system (grey)  or \ac{EMG} (dark green).}
\label{fig:res:squat}
\end{figure}

The \ac{RMSE} value for both \ac{IK} and \ac{ID}, comparing the proposed real-time framework with the offline Mocap-based OpenSim analysis are presented in \cref{tab:full_rmse} for three activities: level ground walking, squat, and STS, performed by two subjects. \cref{fig:res:walk}, Fig.~\ref{fig:res:squat} and \cref{fig:res:sts:ik_id} illustrate the sagittal plane kinematics, kinetics, and estimated muscle activations of selected major lower limb muscles on both sides of the subjects, obtained based on the real-time pipeline and the Mocap-based system, including EMG data. 


\subsubsection{Level ground walking}\label{sec:pipe:walking:res}
Consistent with findings in
\cref{sec:pipe:imu_ik_only:res}, joint kinematics generally showed good agreement with the Mocap-based data, although a noticeable offset was observed in hip flexion/extension. The RMSE ranged from \ang{3} to \ang{8}, with the smallest error observed at the knee joint. 
For \ac{ID}, ankle torque also showed a similar pattern as the reference data. However, a noticeable discrepancy in magnitude were observed at the knee and hip joints, particularly in the push-off phase. Regarding to estimated muscle activations, most muscles showed similar activation patterns to those observed in the measured EMG data, with the exception of the vastus medialis. 


\subsubsection{Squat}\label{res:pipeline:squat}
Clear and symmetric descending and ascending phases can be observed in all joints. In terms of kinematics, the overall patterns show good agreement between the proposed real-time platform and the Mocap-based analysis. However, our platform tends to underestimate the maximum flexion angles, particularly at the ankle and knee where the RMSE ranges from \ang{6} to \ang{14}. For joint kinetics, the real-time joint torque estimation are generally lower than those from the Mocap system, particularly at the knee. Regarding to estimated muscle activations, most muscles show activation pattern similar to those observed in the measured EMG data, with the exception of the tibialis anterior.

\subsubsection{Stand to Sit and Sit to Stand}\label{sec:pipe:sts:res}
Overall, joint kinematics, joint kinetics and estimated muscle activations show good agreement with the reference based on the Mocap system. However, in consistent with the squat, the real-time system still tends to underestimate joint angles, particularly at the hip.

\begin{figure}[!ht]
\centering
 \includegraphics[width=\columnwidth]{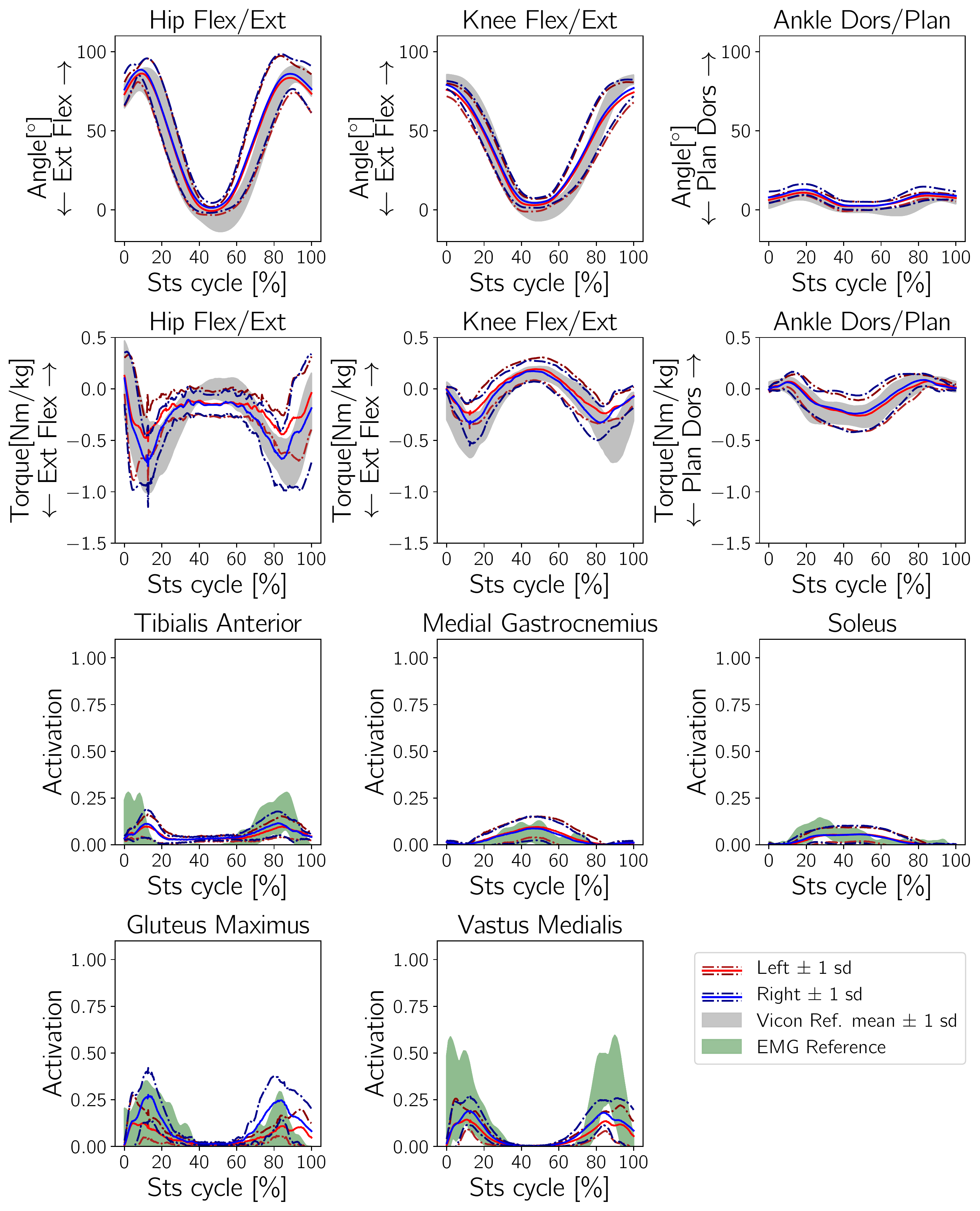}
\caption{Lower limb sagittal plane IK, ID at the ankle, knee and hip, along with muscle activations of tibialis anterior, medial gastrocnemius, soleus, gluteus maximus, and vastus medialis during stand-to-sit and sit-to-stant based on the full-chain real-time pipeline for two subjects. Solid and dashed lines illustrated mean $\pm 1$ \ac{sd} of the left and right sides. Shaded areas indicate reference data obtained based on motion capture system (grey)  or \ac{EMG} (dark green)}
\label{fig:res:sts:ik_id}
\end{figure}

\def \sone {\textbf{{S1}}}
\def \stwo {\textbf{{S2}}}

\begin{table}
 \caption{RMSE of sagittal plane IK and ID for two subjects (S1 and S2) were computed to compare the proposed real-time pipeline output against the offline-processed Mocap-based measurements}
    \label{tab:full_rmse}
\begin{minipage}{\columnwidth}
\begin{tabularx}{\textwidth}{l *{6}{S[round-mode = places,table-column-width = 0.8cm]}  }
\toprule
                                                                             & \multicolumn{2}{c}{\textbf{Walking}}          & \multicolumn{2}{c}{\textbf{Squat}}          & \multicolumn{2}{c}{\textbf{\ac{STS}}}       \\
\cmidrule(l){2-3} \cmidrule(l){4-5} \cmidrule(l){6-7}
                                                                             & \sone                  & \stwo                & \sone                & \stwo                & \sone                & \stwo                \\
\midrule
\multicolumn{7}{l}{\textbf{Angle} [\si{\degree}]} \\ 
                                                           hip       & 7.646992565427052      & 4.563904984722703    & 13.031781407574552   & 9.501050922310478    & 15.094381978814349   & 4.344299425274257    \\
                                                           knee              & 3.979608231377327      & 3.884305741238474    & 13.739592686677524   & 14.143307482804905   & 8.589168342997793    & 6.402471803149622    \\
                                                           ankle             & 6.175763071654982      & 5.119722281869612    & 6.144228072768651    & 9.0890299514821      & 3.3601232936346217   & 3.4988540068830334   \\
\multicolumn{7}{l}{\textbf{Torque} [\si{\newton\meter\per\kilogram}]} \\
                                                           hip       & 0.8311704617633013     & 0.793403919201053    & 0.2175219738896774   & 0.07426546322282838  & 0.20076302551980632  & 0.1756665717612725   \\
                                                           knee              & 0.47560580120897994    & 0.5171355142319838   & 0.18296507398632528  & 0.3485556594558743   & 0.16741930585414885  & 0.13270631747480507  \\
                                                           ankle             & 0.2105214664122145     & 0.1831369084072002   & 0.17045506466786176  & 0.03983509068596096  & 0.0988667070587194   & 0.06333080288017867  \\
\bottomrule
\end{tabularx}
\parnotes
\end{minipage}
\end{table}


\subsection{Loop Times of different components}

\cref{tab:res:SO:threads} presents the loop times for each event across different node components of the pipeline, executed with 4, 6 or 12 threads. Regardless of the number of threads, the longest duration was consistently observed for events 0-1, the synchronization event. The second-longest duration event was the \ac{SO} multithreaded pipeline buffer (event 7-8). However, when running with 12 threads, the synchronization of the \ac{SO} waiting queue (event 7-8) was no longer slower than the \ac{SO} computation itself (event 8-9).

Due to variability in data arrival times, we observed that the platform benefits from running with 12 threads, even though the average processing time for 4 threads was only \qty{22.928316}{\milli\second}.
Finally, when considering the 95\% cumulative probability of receiving valid \ac{SO} results 
the expected computation times were \qty{590.0}{\milli\second}, \qty{480.0}{\milli\second} and \qty{401.0}{\milli\second} for 4, 6 and 12 threads, respectively. With an empirically defined deadline threshold of \qty{0.5}{\second} for the \ac{SO} output, using 12 threads resulted in a delivery rate of 99.1\% of messages being received on time.

\newcommand{\taTimeUnit}[1]{#1}

\begin{table}
\caption{Timings for \ac{ID} and \ac{SO} with 4, 6 and 12 threads\label{tab:res:SO:threads}}
\begin{minipage}{\columnwidth}
\begin{tabularx}{\textwidth}{G{2cm}ZZ *{3}{S[round-mode = places,table-column-width = 2cm]}}

\toprule
                &                       &                           & \multicolumn{3}{c}{Mean latency with N processes (\qty{}{\milli\second})}\\
\textbf{Events\parnote{Events 1-5 took less than \taTimeUnit{0.02}\qty{}{\milli\second} and are therefore omitted from this table.}}	& \textbf{End event}    & \textbf{Reference event}	& \textbf{N = 4}& \textbf{N = 6}& \textbf{N = 12}\\
\midrule
0-1& \variableEventB  	&  \variableEventA 	 &  \taTimeUnit{260.602446}  & \taTimeUnit{260.627500}   &  \taTimeUnit{260.692567} \\
\midrule

5-6& \variableEventG  	&  \variableEventF 	 &  \taTimeUnit{0.113452}    & \taTimeUnit{0.099791}     &  \taTimeUnit{0.102790}\\
6-7& \variableEventH  	&  \variableEventG 	 &  \taTimeUnit{23.699341}   & \taTimeUnit{19.249667}    &  \taTimeUnit{18.281034}\\
7-8& \variableEventI  	&  \variableEventH 	 &  \taTimeUnit{55.889934}   & \taTimeUnit{21.855266}    &  \taTimeUnit{2.911814}\\
8-9& \variableEventJ  	&  \variableEventI 	 &  \taTimeUnit{22.928316}   & \taTimeUnit{26.580842}    &  \taTimeUnit{28.411721}\\
\midrule
All & all 	&  	                             &  \taTimeUnit{363.257385}  & \taTimeUnit{328.442392}   &  \taTimeUnit{310.424078}\\
\textbf{95\% Latency} & all 	&  	                             &  {590.0}  & {480.0}   &  {401.0}\\
\bottomrule
\end{tabularx}
\parnotes
\end{minipage}

\end{table}

\section{Discussion}


\label{sec:discussion}

In this paper, we presented a novel proof-of-concept integration framework that enables real-time estimation of musculoskeletal model-based joint angles, torques, and muscle activation. The integrated framework leveraged the advantages of a common musculoskeletal modeling framework OpenSim and the robotic operating system, supporting a sensor-agnostic, extensible, open-source, and wearable setup.  Any sensors, such as \ac{IMU}, camera or force-sensing insole whose data can be read from a Linux system may be used integrated. As part of bench testing, we evaluated the real-time \ac{IK} of the upper body movement (thorax, arm, and elbow) using \ac{AR} markers, and lower body movement (ankle, knee, hip, and pelvis) during level-ground walking using \ac{IMU}s. We further demonstrated the full-chain real-time computation pipeline including lower body \ac{IK}, \ac{ID}, and muscle activation during walking and other daily activities using \ac{IMU}s and pressure insoles. To the best of our knowledge, it is the first real-time framework of its kind with an integration of multiple heterogeneous wearable sensors, \ac{ROS} and OpenSim. The proposed system has strong potential for various application scenarios such as quantitative movement assessment, rehabilitation, and human-centered robotic control. Its modular design improves extensibility and broadens sensor compatibility compared to existing real-time frameworks\cite{slade_open-source_2021,stanev_real-time_2021}. As such, our work opens new avenues for real-time musculoskeletal simulation and provides a foundational platform that can be extended with additional sensor modalities, activity scenarios, and model types.

Given the time-sensitive nature of data processing required for real-time musculoskeletal simulations, \ac{ROS} serves as a suitable candidate as a middleware component. 
However, integration efforts between \ac{ROS} and biomechanical simulation platform such as OpenSim have been limited. Despite this, there are several structural analogies between the OpenSim pipeline and the general architecture of \ac{ROS}, particularly in model construction, \ac{IK} and \ac{ID} handling, and the use of debugging tools. 
One of the key strengths of \ac{ROS} is its publisher-subscriber communications model, which enforces clear, modular interfaces. Data such as angles obtained from \ac{IK} or forces from \ac{ID} are transmitted using standardized messages, enabling seamless integration with other software components, such as model and sensor data visualization. In addition, the \ac{ROS} scheduling framework supports multithreading execution, as demonstrated by our implementation of a multithreading \ac{SO} node. This capability allows researchers to focus on the algorithms development rather than dealing with interprocess communication. In addition, the use of a containerized system via Docker facilitates both replicability and extensibility of the framework. 
However, it is worth noting the communications overhead introduced by having multiple successive nodes. This contributes to increased timing variability of the system, which can either add latency to downstream notes or reduce their effective frame rates. Based on our loop time analysis, this effect was particularly evident in the \ac{SO} output. Although the \ac{SO} computation itself requires only 20-30 ms on average, it required an additional 190 ms to deliver 99\% of frames on time. We believe that this level of delay is acceptable for applications such as exoskeleton control during slow to normal walking, or in real-time rehabilitation scenario. However, for fast activities such as running, this delay may become a limiting factor.

The strict use of low-cost wearable sensors enables the widespread use of this technology in out-of-the-lab settings and across various fields such as ergonomics, rehabilitation, and healthcare monitoring. In this study, we evaluated two types of motion-tracking input methods: image-based visual tracking with \ac{AR} markers, and \ac{IMU}-based inertial tracking. Similar to the single-camera set-up by Nagymáté et al.~\cite{nagymate_affordable_2019}, joint angles were estimated by tracking the poses of the \ac{AR} markers, based on the locations of the virtual anatomical landmarks and orientations of the segments. We found that \ac{AR} marker-based upper body joint angle closely matched with those estimated by \ac{IMU} during simple planar movements. However, accurate pose estimation using \ac{AR} markers requires markers to be sufficiently large and mounted on rigid surfaces. The detection accuracy is also highly sensitive to image quality. A camera with high resolution and high frame rate is often required for capturing faster motions. Marker occlusion might be a limiting factor in complex 3D motions with larger motion volumes. In our setup, for example, the chest marker became undetectable when the subject had more than \ang{20}  of trunk flexion. However, we believe that the limitation can be addressed in future implementations of multiple markers per segment at different orientations and/or incorporating additional cameras to improve tracking robustness.

\ac{IMU}-based inertial motion tracking is among the most widely studied wearable motion tracking methods for measuring joint kinematics in large cohorts and the natural environment. To address common limitations such as long-term drift, magnetic interference, and measurement inconsistency, researchers have developed various sensor fusion methods, most notably in the field of visual-inertial odometry~\cite{zhang_tutorial_2018,forster_-manifold_2017}, machine learning~\cite{zhang_fusing_2020,malleson_real-time_2020,gilbert_fusing_2019} and biomechanical models-constraint methods~\cite{slade_open-source_2021,al_borno_opensense_2022}. However, most of these methods require off-line post-processing or rely on proprietary software (e.g., Xsens MVN), which limits interoperability with other systems~\cite{slade_open-source_2021}.
When compared to the open-source Mocap-based reference, our \ac{IMU}-based estimates of lower extremity joint angles showed good agreement during level-ground walking except for a clear offset in hip flexion/extension and pelvic tilt present on \cref{fig:ik_gait}. Such discrepancies have been previously reported in literature~\cite{lathrop_comparative_2011,roelker_interpreting_2017,wouda_validity_2018,opensim_gait_2024} and may depend on the musculoskeletal model used. Notably, these offsets were eliminated in comparisons using our own simultaneously acquired reference data, where the same musculoskeletal model was applied in both cases (\cref{fig:res:sts:ik_id}). Lower extremity angles in other motions, i.e., squat and \ac{STS}, generally show good agreement. However, we observed that the underestimated peak ankle and knee angles during squat may have been constrained by the distal placement of the tibial \acp{IMU}, which likely resulted in sensor displacement during the deepest portion of the movement. A similar issue may have affected the pelvis IMU during \ac{STS} for one subject, potentially contributing to the high RMSE observed in the hip angle. These sensor placements induced artifacts in turn affected downstream analysis such as ankle torque estimation and tibialis anterior activation estimation. Finally, the use of flexible straps for \acp{IMU} mounting as opposed to rigid plates used in some other studies, may have introduced additional motion artifacts, contributing to a slightly higher average \ac{RMSE} values compared to results reported by~\cite{al_borno_opensense_2022}.

Wireless pressure insoles overcome the limitation of costly and stationary force plates, which restrict  \ac{GRF}s and \ac{COP} to short recordings with limited steps in controlled laboratory  environment. We noticed that the current insole setup provides reasonable accurate lower limb joint torque estimations in the \ac{STS} and squat when both feet remain fully in contact with the ground.  These results suggest that current real-time framework with IMU and pressure insole is well suited for analyzing quasi-static or symmetrical lower-body activities. This has promising implications for a range of practical applications such as strength and balance exercises during rehabilitation, joint loading analysis during strength training, and ergonomics assessment to lifting techniques. 

It worth noting that current commercial wireless pressure insoles come along with inherent limitations, e.g. only vertical component of the \ac{GRF} measured and require a non-trivial \ac{COP} transformation from a local coordinate system to a global coordinate system. During level-ground walking, ankle joint torque in the sagittal plane showed a similar pattern compared to the reference data. However, both the knee and hip exhibited higher torque peaks during pre-swing phase. This was primarily attributed to an unrealistic high second peak in the \ac{GRF} measured by the pressure insole, inaccuracies in \ac{GRF} orientation due to the absence of shear force in the anterior-posterior direction, and less accurate \ac{COP} transformation. Previous studies using OpenGo have reported that peak forces measured by pressure insoles can up to 44\% lower than those recorded by force plates for both walking and running \cite{stoggl_validation_2017}. Based on our observation, the normal force measured by the insole tends to be underestimated at heel strike and overestimated at push-off. Depending on modeling assumptions,  whether the foot segment or ground is considered rigid, the normal force may be applied either always normal to the foot segment or to the ground, respectively. However, neither assumption fully captures the true \ac{GRF} vector during walking. Assuming the force is normal to the foot can lead to  underestimation of normal forces and overestimate shear forces during loading response and pre-swing, resulting in an unrealistic larger moment arm, particularly at the more proximal joints such as the knee and hip. After a preliminary examination, we considered it more reasonable to assume a rigid ground and apply the insole measured force normal to the ground. Nevertheless, there is still a profound inaccuracy in knee and hip joint torque estimations during level-ground walking. 
The limitations of current pressure insole technology constrain of our framework to a broader range of activities. Potential solutions include augmenting insole data with analytical or machine learning-based \ac{GRF} estimation. 

A major limitation of the current study was that the small sample size, that restricts our ability to draw conclusions about the precision of the proposed real-time framework. Future work will focus on enhancing the robustness of the framework and evaluating the accuracy of \ac{IK}, \ac{ID}, and \ac{SO} estimations in a larger cohort.

\section{Conclusion}

We proposed a novel integration of \ac{ROS} with \ac{opensimrt}, resulting in a framework for real-time estimation of joint angles, torques, and muscle activations driven by wearable sensor and musculoskeletal modeling. As a proof-of-concept, we evaluated the proposed framework on healthy subjects with a wearable setup of either \ac{IMU} or \ac{AR} markers, along with pressure insoles. While further validation is needed, we observed biologically plausible results in the tested activities, especially in the 3D joint kinematics, ankle joint torque, and muscle activations. Additionally, we also showed that the delays introduced by sensors, socket communications, and multiple chained optimizers did not hinder real-time performance for evaluating gait and other daily activities. We believe that our proposed framework represents a significant step toward more accessible and real-time human movement analysis and shows great potential for advancing technologies in biofeedback-driven rehabilitation and exoskeleton control.

\section*{Acknowledgment}
We thank student Mårten Norman for assisting with data collection. 

\bibliography{references.bib}

\end{document}